\documentclass[manuscript]{acmart} 

\AtBeginDocument{%
  }

\renewcommand\footnotetextcopyrightpermission[1]{} % Suppresses the footnote with conference information
\acmConference{}
\acmYear{}
\acmDOI{}
\acmISBN{}
\acmBooktitle{}

\usepackage{geometry}
\usepackage{tabularx}
\usepackage{booktabs}
\usepackage{array}
\usepackage{multirow}
\usepackage{caption}
\usepackage{pdflscape}
\usepackage{pgfplots}
\usepgfplotslibrary{groupplots}
\usepackage{afterpage}
\usepackage{arydshln}
\usepackage{graphicx}
\usepackage{datatool}
\usepackage{tikz}
\pgfplotsset{compat=1.17}
\usepackage{subcaption}
\usetikzlibrary{patterns}
\usepackage[title]{appendix}

\begin{document}

\title{A Comparative Analysis of Instruction Fine-Tuning LLMs for Financial Text Classification}

%% define my own color to use in the coloring of the content

\definecolor{mygreen}{HTML}{5F9EA0} % Define 'mygreen' as #5F9EA0

\author{SorourAlsadat Fatemi}
\affiliation{%
  \institution{University of Illinois at Chicago}
  \city{Chicago}
  \country{USA}}
\email{sfatem6@uic.edu}
\orcid{\textcolor{red}{1234-5678-9012}}

\author{Yuheng Hu}
\affiliation{%
  \institution{University of Illinois at Chicago}
  \city{Hekla}
  \country{USA}}
\email{yuhenghu@uic.edu}

\author{Maryam Mousavi}
\email{m.mousavi@asu.edu}
\orcid{\textcolor{red}{1234-5678-9012}}
\affiliation{%
  \institution{Arizona State University}
  \city{Tempe}
  \state{AZ}
  \country{USA}
}

% %%
% %% By default, the full list of authors will be used in the page
% %% headers. Often, this list is too long, and will overlap
% %% other information printed in the page headers. This command allows
% %% the author to define a more concise list
% %% of authors' names for this purpose.
\renewcommand{\shortauthors}{Fameti et al.}

\begin{abstract}

Large Language Models (LLMs) have demonstrated impressive capabilities across diverse Natural Language Processing (NLP) tasks, including language understanding, reasoning, and generation. However, general-domain LLMs often struggle with financial tasks due to the technical and specialized nature of financial texts. This study investigates the efficacy of instruction fine-tuning smaller-scale LLMs, including Mistral-7B, Llama3-8B, and Phi3-mini, to enhance their performance in financial text classification tasks. We fine-tuned both instruction-tuned and base models across four financial classification tasks, achieving significant improvements in task-specific performance. Furthermore, we evaluated the zero-shot capabilities of these fine-tuned models on three unseen complex financial tasks including argument classification, deal completeness classification and causal classification. Our results indicate while base model fine-tuning led to greater degradation, instruction-tuned models maintained more robust performance. To address this degradation, we employed model merging techniques, integrating single-task domain-specific fine-tuned models with the base model. Using this merging method resulted in significant enhancements in zero-shot performance, even exceeding the original model's accuracy on certain datasets. Our findings underscore the effectiveness of instruction fine-tuning and model merging for adapting LLMs to specialized financial text classification tasks.

\end{abstract}

\maketitle

\section{Introduction}

Large Language Models (LLMs) have revolutionized Natural Language Processing (NLP) by demonstrating exceptional capabilities in understanding, reasoning, and generating human-like text across various general-domain tasks. Prominent models like ChatGPT \cite{chatgpt2023optimizing} and GPT-4 \cite{openai2023gpt} have shown impressive versatility in following general human instructions and handling various tasks such as question answering \cite{chen2023distinguish}, machine translation \cite{zhu2023multilingual}, information extraction \cite{dunn2022structured}, and grammar correction \cite{omelianchuk2024pillars}. This broad proficiency has led to growing interest in leveraging LLMs for industry-specific applications including medicine and finance. For instance, Med-PaLM 2 \cite{singhal2023towards} has been adapted for medical domains to provide accurate responses to medical queries, while Bloomberg's financial LLM \cite{wu2023bloomberggpt} supports various NLP tasks within the financial sector. \\

In the finance industry, beyond quantitative data typically analyzed, the sentiment and tone of financial reports, earnings calls, news articles, and social media posts significantly influence investor decisions. Therefore, extracting and analyzing relevant textual information is critical for informed investment strategies and decision-making \cite{gross2011machines, ederington1993markets}. Despite the advancements of general-domain LLMs, they often fall short when applied to specialized fields like finance due to complex terminologies and intricate concepts. This underscores a need for domain-specific adaptations to fully realize the potential of LLMs in financial applications.\\

Existing research highlights the success of LLMs in financial tasks, such as predicting stock price movements \cite{xie2023wall, lopez2023can, li2023chatgpt} and performing advanced financial text analytics \cite{xie2024pixiu, fatouros2023transforming, xie2024finben, fatemi2023comparative}. However, significant challenges remain, particularly in tasks like financial relation extraction and numerical reasoning, which are crucial for making well-informed decisions \cite{li2023chatgpt, xie2024finben}. The efficient market hypothesis \cite{malkiel2011efficient} further underscores the importance of linking public information with stock returns, emphasizing the need for sophisticated analysis tools \cite{gross2011machines, ederington1993markets}. % Recently, Large Language Models (LLMs) have demonstrated potential in automating labor-intensive aspects of financial analysis due to their advanced natural language understanding, reasoning, and generation capabilities \cite{nori2023capabilities,bubeck2023sparks, chen2023distinguish, adlakha2024evaluating}.
\\

To address these limitations and enhance the capabilities of LLMs for specific domains, existing methods can be broadly classified into three main approaches: in-context learning, training models from scratch on domain-specific and general data, and fine-tuning existing models using supervised datasets. In-context learning, where LLMs generate results based on a few demonstration examples \cite{li2023chatgpt}, can be costly, slow in inference, and limited by the model's context window \cite{bertsch2024context}. Moreover, these models can be sensitive to the quality and variability of the provided examples \cite{islam2023financebench}. Training models from scratch, while effective, demands significant computational resources and vast domain-specific and general dataset. Fine-tuning existing models is promising alternative but faces challenges such as the need for high-quality datasets and the risk of catastrophic forgetting, where the model's ability to perform general tasks degrades after domain-specific fine-tuning. Techniques like model merging can help mitigate these effects.\\

Previous studies have demonstrated that instruction fine-tuning smaller open-source language models can significantly enhance their performance on domain-specific tasks across various fields, such as law and medicine. In many cases, these models have outperformed the zero-shot performance of proprietary LLMs and other state-of-the-art models \cite{zhang2023multi, zhang2023instruct, huang2023lawyer, li2023beginner, mousavi2022effective}. Within the finance domain, one study fine-tuned the Llama2 model for sentiment analysis, outperforming the FinBERT \cite{zhang2023enhancing}. Another study fine-tuned Llama2-7B and Llama2-13B models on a variety of financial tasks, including sentiment analysis, relation extraction, question answering, and stock market prediction \cite{xie2024pixiu}. However, much of the existing research has focused primarily on fine-tuning models from the Llama family. To address this gap, we explore fine-tuning other powerful, smaller LLMs, specifically Mistral-7B, Phi-3, and Llama2-8B, across four representative financial text classification tasks: sentiment analysis, news headline classification, relation extraction, and hawkish-dovish classification.\\

One of the main challenges with fine-tuning LLMs for domain-specific tasks is the degradation of their zero-shot performance on unseen tasks \cite{zhang2023instruction}. To mitigate this, previous studies have incorporated general instruction data into domain-specific training datasets or augmented the data to reduce model sensitivity to semantically similar prompts \cite{zhang2023multi, liu2023goat, wang2023huatuo}. While this approach improves generalizability, it also increases computation cost and training time.

\subsection{Our approach and Contributions}

     In our study, we aimed to address these challenges and improve the performance of smaller LLMs on specialized financial tasks by focusing on in-context learning, fine-tuning and model merging techniques. By leveraging these approaches, we aim to improve the performance of small LLMs in specialized financial tasks while maintaining their general capabilities on unseen tasks. We explored the potential of \textbf{Mistral-7B}, \textbf{Phi-3}, and \textbf{Llama2-8B}, which are powerful yet resource-efficient models, across seven representative financial text classification tasks: \textbf{sentiment analysis}, \textbf{news headline classification}, \textbf{relation extraction}, \textbf{hawkish-dovish classification}, \textbf{argument unit classification}, \textbf{deal completion classification} and \textbf{causal classification}. These models were selected to address the challenge of adapting LLMs to complex, domain-specific tasks while managing computational costs. By concentrating on smaller models, we aim to provide a more scalable solution that balances performance with efficiency in the financial sector.\\
    
    We first assessed the in-context learning capabilities of selected models on the specified tasks. We experimented with one-shot, five-shot and ten-shot examples and compared them against zero-shot performance of the models. The results shows some improvement in the model prediction when providing the in-context examples. However, after a certain number of examples, increasing the number of demonstrations led to performance degradation in small instruct models.\\
    
    Next, we fine-tuned the models to improve task performance while maintaining the generalization capabilities of the models. Both base and instruction-tuned models were fine-tuned using the same set of finance-specific training data. Our approach differs from previous work by focusing on instruction-tuned models that had already been exposed to a diverse range of tasks, reducing the need for extensive general datasets. The results show that fine-tuning both base and instruct models leads to substantial improvements over zero-shot performance, including when compared to proprietary models such as GPT-4. In particular, the models excelled in tasks like relation extraction and hawkish-dovish classification, which are typically underrepresented in the pre-training data of most LLMs. We then evaluated the performance of fine-tuned models on three different unseen tasks and we observed that base fine-tuned models exhibited greater performance degradation compared to instruct fine-tuned models. These findings indicate that instruct fine-tuned models provide a more robust foundation for further fine-tuning on domain-specific tasks.\\
    
    To further address the performance degradation on unseen tasks, we leveraged the merging techniques using \textbf{MergeKit framework}. This allowed us to merge single-task fine-tuned models with the vanilla instruction model, helping to preserve the models' zero-shot generalization abilities while enhancing their task-specific performance. Notably, the Mistral-7B model demonstrated results that were either on par with, or surpassed, its original zero-shot performance, highlighting the effectiveness of model merging techniques in maintaining robust performance across tasks.\\
    
    Our key contributions are as follows:
    \begin{itemize}
        \item We experimented with in-context learning of small LLMs, including Llama3-8B, Mistral-7B, and Phi-3-mini, on four financial domain datasets. Our findings indicate that increasing the number of examples degrades the performance of small instruct models. 
        \item We fine-tuned smaller models such as Mistral-7B, Phi-3, and Llama2-8B on four key financial text classification tasks, showcasing the feasibility of using smaller, more efficient models for domain-specific financial tasks.
        \item We demonstrated significant performance improvements across all models by fine-tuning both base and instruct models, with the instruct models proving more robust in handling complex financial tasks.
        \item We introduced model merging techniques via the MergeKit framework, which learned the domain specific knowledge effectively, while mitigated the typical degradation of zero-shot performance on unseen tasks, notably improving results for the Mistral-7B model.
        \item We highlight the advantages of fine-tuning smaller LLMs on specialized financial datasets, achieving strong performance without the extensive resource requirements typically associated with larger models.
    
    \end{itemize}

\section{Related Work}
This section reviews recent studies that apply Large Language Models (LLMs) to domain-specific tasks. Prior to the rise of LLM-focused research, many studies concentrated on utilizing BERT, an encoder-only model, for improving performance through fine-tuning, developing self-attentive mechanisms, and incorporating multiple lexical knowledge sources to better capture semantic context, specially in the finance domain \cite{du2023incorporating, xiang2022semantic, yang2020finbert, shah2022flue, mousavi2021stif}. However, this paper focuses on instruction fine-tuning, which leverages decoder-only LLM models optimized for generating text based on preceding context.

\subsection{Training from Scratch}
Training domain-specific language models from scratch remains a straightforward method for achieving domain adaptation. BloombergGPT is an early example of large language models built specifically for the financial domain. It was trained on a blend of financial and general corpora, and showing high performance on financial tasks \cite{wu2023bloomberggpt}. Similarly, other studies transformed large-scale raw domain-specific corpora into reading comprehension task, allowing LLMs to acquire domain knowledge and improve prompting capabilities \cite{cheng2023adapting}. However, training from scratch demands substantial computational resources and large dataset, making it less practical than other methods such as In-Context Learning (ICL) or fine-tuning, which require fewer resources \cite{yang2023fingpt}.

\subsection{In-context Learning}
% The research community has seen numerous studies focusing on the analysis and enhancement of demonstrations in ICL \cite{wang2023chatgpt,sun2023text, zhang2023sentiment}.
Scaling model size and leveraging extensive pre-training data have unlocked emergent capabilities in LLMs, such as In-Context Learning (ICL) \cite{wei2023larger, chowdhery2023palm, brown2020language}. This allows models to learn from a few in-context provided instructed examples without requiring full retraining \cite{ouyang2022training}. Proprietary models like ChatGPT and GPT-4 have demonstrated strong performance across various general tasks \cite{sun2023text, zhang2023sentiment, zhong2023can}. However, when applied to specialized domains like finance and social media, their performance often lags behind that of fine-tuned models like BERT, particularly for tasks requiring deeper semantic understanding \cite{wang2023chatgpt}. Increasing the number of in-context examples can also lead to performance degradation in certain domain-specific tasks \cite{zhang2023sentiment}.\\

In finance, Li et al. \cite{li2023chatgpt} examined the performance of ChatGPT and GPT-4 across several financial text analytics tasks. Another study enhanced financial sentiment analysis using semi-supervised learning and Chain-of-Thought reasoning to improve accuracy with minimal prompting \cite{deng2023llms}. Although these models outperform some domain-specific models in certain tasks, they struggle with tasks like named-entity recognition and headline classification, which require deeper semantic analysis. These limitations motivate our research into the performance of open-source instruction-tuned models in the finance domain.

\subsection{Instruction Fine-tuning}

Instruction fine-tuning is a resource-efficient approach for adapting LLMs to specific tasks. It allows models to adjust to domain-specific needs without the need for extensive retraining \cite{zhang2023instruction, ouyang2022training}. This technique has proven effective across domains like medicine, law, and social sciences, where fine-tuned models often outperform general-purpose models like GPT-4 \cite{li2023beginner, liu2023goat, zhang2023multi, huang2023lawyer, yunxiang2023chatdoctor, chiang2023vicuna}.\\

In the finance domain, several studies have applied instruction fine-tuning to models like LLaMA using financial datasets. Xie et al. \cite{xie2024pixiu} fine-tuned LLaMA on 136k task-specific instruction samples, including sentiment analysis, named-entity recognition, question answering, and stock movement prediction, achieving results comparable to GPT-4. Yang et al. \cite{yang2023fingpt} developed an end-to-end framework for training and deploying FinLLMs in the finance sector using Low-Rank Adaptation (LoRA) to fine-tune LLaMA and ChatGLM models using 50k data points from news articles, social media posts, Sec fillings and stock movement predictions. Zhang et al. \cite{zhang2023instruct} also applied fine-tuning to financial classification tasks, focusing on sentiment analysis.\\

However, these studies have mainly focused on fine-tuning the LLaMA model, often overlooking other smaller, open-source LLMs. Moreover, these studies focus on fine-tuning base models which typically require large supervised instruct-created datasets \cite{taori2023stanford}. To overcome data scarcity limitations, Xie et al. \cite{xie2024pixiu} augmented the data by multiple instructions per sample to enhance training, though this increased computational costs. Recently released instruction-tuned models, already trained on diverse tasks, may offer a more efficient alternative by bypassing the need for such extensive data augmentation. To address this research gap, our study fine-tunes both base and instruction-tuned variants of three robust small LLMs (Llama3, Mistral-7B, and Phi-3) on four financial classification tasks using a consistent amount of training data.

\subsection{Merging Models}

In addition, previous studies show that fine-tuning on domain-specific data can degrade performance on unseen tasks. To address this, some researchers in non-financial domains have incorporated both generic and domain-specific instruction data, but this adds significant computational costs \cite{wang2023huatuo, wang2023instructuie}. In the finance domain, \cite{zhang2023instruct, xie2024pixiu} have shown that fine-tuning improves performance on trained datasets, however, these studies have not examined performance on unseen finance-related tasks. To fill this gap, we evaluate the performance of fine-tuned models on three unseen finance-related datasets. In contrast, our study takes a more efficient approach: rather than augmenting training data with generic datasets, we fine-tuned instruction models on each task individually and applied the MergeKit framework to combine the single-task fine-tuned models with the vanilla instruction model, thus preserving performance on unseen tasks while minimizing computational overhead.\\

\begin{figure}[htbp] % Floating location
    \centering
    \includegraphics[width=\textwidth]{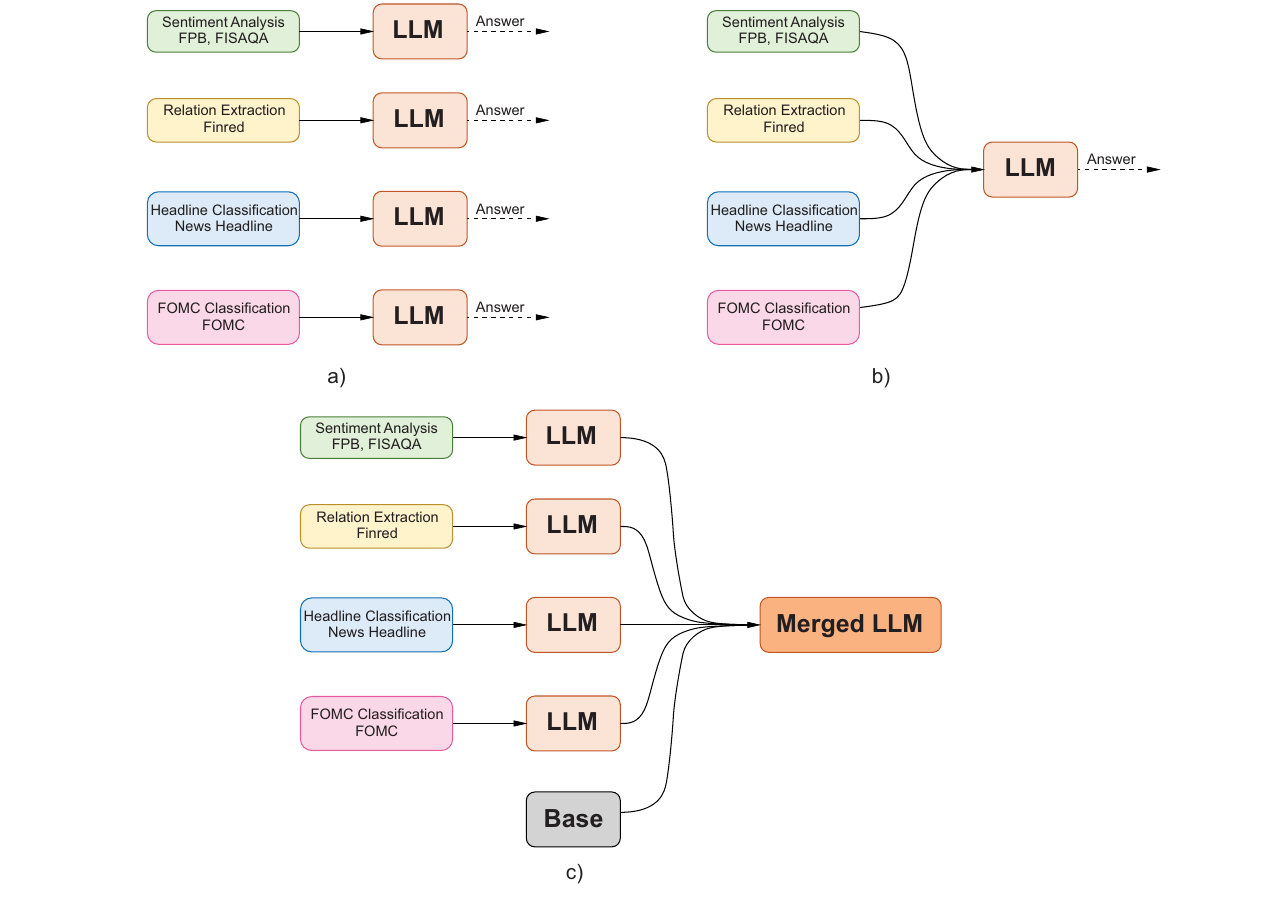}
    \caption{The three approaches used in this work a) single task fine-tuning,  b)Multi-task fine-tuning c)Merging with vanilla models}
    \label{fig:flowchart}
\end{figure}

\begin{table}[ht]
    \centering
    \begin{tabularx}{\textwidth}{X X X c c X}
        \toprule
        \textbf{Dataset} & \textbf{Used for} & \textbf{Task} & \textbf{\# Labels} & \textbf{Data size} & \textbf{Example labels} \\ 
        \midrule
        FPB & Train, test of ICL, FT & Sentiment analysis & 3 & 4,845 & negative, neutral, positive \\
        FiQA-SA & Train, test of ICL, FT & Sentiment analysis & 3 & 1,173 & negative, neutral, positive \\
        Headline-Dir & Train, test of ICL, FT & News headline classification & 3 & 9,277 & up, down, stable \\
        FinRED & Train, test of ICL, FT & Relation extraction & 29 & 1,070 & employer, industry, owner \\
        FOMC & Train, test of ICL, FT & Hawkish-dovish classification & 3 & 496 & hawkish, neutral, dovish \\
        FinArg-AUC-T1 & Evaluation on unseen data & Argument unit classification & 2 & 969 & claim, premise \\
        M\&A & Evaluation on unseen data & Deal completeness classification & 2 & 500 & complete, rumour\\
        FinCausual`20-T1 & Evaluation on unseen data & Causal classification & 2 & 800 & causal, noise \\ 
        \bottomrule
        
    \end{tabularx}
    \caption{Dataset size and number of labels of datasets used in our experiments}
    \label{table:1}
\end{table}

% \begin{table}[ht]
%     \centering
%     \begin{tabular}{@{}lclccc@{}}
%         \toprule
%         Dataset & Used for & Task & \# Labels & Data size & Example labels \\ 
%         \midrule
%         FPB & Train, test of ICL, FT & sentiment analysis & 3 & 22.7  & negative, neutral, positive \\
%         FiQA-SA & Train, test of ICL, FT & sentiment analysis & 3 & 23.7  & negative, neutral, positive \\
%         Headlines-Dir & Train, test of ICL, FT & news headline classification & 3 & 20.7  & up, down, stable \\
%         FinRED & Train, test of ICL, FT & relation extraction & 29 & 27.4  & employer, industry, owner \\
%         FOMC  & Train, test of ICL, FT & hawkish-dovish classification & 3 & 22.3  & hawkish, neutral, dovish \\
%         FinArg AUC Task1 & Train, test of ICL, FT & argument unit classification & 2 & 22.3 & claim, premise \\
%         M\&A  & Evaluation on unseen data & deal completeness classification & ? & 22.3  &  ?\\
%         \textcolor{red}{FinCausual 2020 Task1} & Evaluation on unseen data & causal classification & 2 & 22.3  & causal, noise \\ \bottomrule
%     \end{tabular}
%     \caption{The datasets we consider in this work span diverse label spaces and domains. The average demonstration length is the average combined length of input, output, and formatting tokens per demonstration provided in the context window.}
% \end{table}

\section{Experiment Setup}
In this section, we provide a detailed overview of the experimental setup used for both in-context learning (ICL) and fine-tuning experiments. This setup includes the selection of models, the configuration of training parameters, and the choice of datasets, all carefully designed to assess the effectiveness of various approaches in adapting large language models to domain-specific financial tasks. By standardizing these conditions, we aim to ensure a fair comparison between different techniques, ultimately highlighting the strengths and limitations of each method in enhancing model performance across specialized financial text classification tasks.

\subsection{Models}
We conduct ICL and fine-tuning experiments using various open-source small models to examine the effects of different model sizes and pre-training datasets. All models employed in this study utilize a decoder-only transformer architecture. We selected three open-source models because they are among the most powerful small-size open-source models available. We fine-tune both the instruct and base models for the Llama3 and Mistral models, and only the instruct model for the Phi3 model since the base model was not available. It should be noted that ICL experiments were exclusively conducted on the instruct models.

\begin{enumerate}
    \item \textbf{Llama3-8B}\cite{llama3}: This latest version from the Llama family is pre-trained on 15 trillion tokens. Compared to its predecessor (e.g. Llama-2), it features a larger tokenizer and a higher-quality training dataset obtained through an enhanced data-filtering pipeline. The instruction-fine-tuned variant of this model has undergone a combination of supervised fine-tuning (SFT), rejection sampling, proximal policy optimization (PPO), and direct preference optimization (DPO).
    \item \textbf{Mistral-7B}\cite{jiang2023mistral}: The smallest model in the Mistral family, this 7-billion parameter model is fine-tuned to achieve performance surpassing that of all other models of similar size on the MT-Bench benchmark.
    \item \textbf{Phi-3-mini-Instruct}\cite{abdin2024phi}: This model, with 3.8 billion parameters, is pre-trained on 3.3 trillion tokens. We utilized the instruction-fine-tuned version, which incorporates post-training processes like supervised fine-tuning and direct preference optimization to enhance instruction-following capabilities.
\end{enumerate}

\subsection{Tasks and Datasets}
We conduct ICL and fine-tuning experiments on four financial text classification tasks: sentiment analysis, relation extraction, news headline classification, and hawkish-dovish classification \footnote{Some of these dataset were obtained from the HuggingFace\url{https://huggingface.co/TheFinAI} and \url{https://huggingface.co/FinGPT}}. Details of the dataset are provided in Table \ref{table:1}, and splitting methods and statistics are presented in Table \ref{table:4} in appendix \ref{subsec:appendix:a}.

\textbf{Sentiment Analysis} is a critical tool in finance, used to predict market trends and investment behavior by analyzing news and social media data \cite{mishev2020evaluation, li2023chatgpt}. Timely extraction of sentiment from these sources is crucial for decision-making by traders and investors. Following BloombergGPT's method \cite{wu2023bloomberggpt}, we use two sentiment datasets, reserving 20\% of the labeled data for testing.
\begin{enumerate}
    \item Financial PhraseBank \cite{malo2014good}: Financial PhraseBank \cite{malo2014good}: This dataset contains sentiment classifications (positive, negative, neutral) derived from financial news, annotated by 5-8 individuals. We focus on the subset with at least 50\% agreement among annotators.
    \item FiQA Sentiment Analysis \cite{yang2018financial}: This dataset comprises 961 samples, each annotated with one of three labels: positive, neutral, or negative.
\end{enumerate}

{\textbf{News Headlines Classification} focuses on extracting actionable information from news beyond basic sentiment analysis, which is valuable for investors, policymakers, and market practitioners. Utilizing the Headlines dataset \cite{sinha2021impact} with 11,412 annotated news headlines about "gold" from 2000 to 2019, we extract additional dimensions like price movements. Specifically, we focus on a subset converting it into a three-class dataset identifying gold prices as Up, Down, or Stable.

\textbf{Hawkish-Dovish Classification} involves categorizing Federal Open Market Committee (FOMC) monetary policy statements as either hawkish or dovish. This classification is significant due to its impact on financial market returns. Conventional sentiment analysis models, which typically categorize text as positive or negative, struggle to capture the nuanced policy stance in these texts accurately. As mentioned in the study, for example, a sentence containing the word "increase" could be either dovish or hawkish depending on context, without necessarily conveying negativity. To address this challenge, we utilized a dataset where FOMC statements were annotated as Hawkish, Dovish, or Neutral, following the splitting method outlined in the original study \cite{shah2023trillion}.

\textbf{Relation extraction} plays a crucial role in financial text analysis by identifying relationships between entities, which supports tasks like knowledge graph creation, question answering, and semantic search. FinRed dataset focuses on this task and identifys relationships in financial news and earnings transcripts. We transform the dataset into a classification task that identifies the relationship between two entities in a sentence, categorizing 29 relation classes such as "subsidiary" and "manufacturer". We reserve 20\% of the data for testing purposes \cite{sharma2022finred}.

\subsection{Unseen Tasks and Datasets}
To assess the generalizability of our fine-tuned models, we evaluated their performance on three unseen tasks. Details of the dataset are provided in Table \ref{table:1}, and in Table \ref{table:4} in appendix \ref{subsec:appendix:a} \\

\textbf{Argument Unit Classification} goes beyond sentiment analysis by exploring the detailed elements of market dynamics and financial events. It entails identifying and categorizing specific units or segments of arguments within earnings conference call data. This classification is fundamental for a detailed breakdown of financial narratives, facilitating better comprehension and analysis. We used FinArg AUC dataset to classify sentences as either claims or premises, facilitating a more nuanced analysis of financial narratives \cite{sy2023fine}.\\

\textbf{Causal Classification} identifies implicit causal relationships within financial documents. Using the SC dataset, sentences from financial news and Securities and Exchange Commission (SEC) filings were classified as either causal or noise, highlighting the underlying causes that influence market trends \cite{mariko2020financial}.\\

\textbf{Deal Completeness Classification} focuses on determining the status of mergers and acquisitions (M\&A) events, distinguishing between completed deals and ongoing rumors. The dataset includes news articles and tweets related to M\&A events, with each instance describing a potential deal between an acquirer and a target company, including IPO rumors. Each instance is categorized as either complete (successful deal) or rumor (no deal materialized). This task is crucial due to the impact of M\&A deal rumors on the share price volatility of target firms, influencing cumulative abnormal returns \cite{yang2020generating}.

\subsection{Methodology}
In this section, we outline the methodologies employed in our study, beginning with a brief introduction to the setup of instruction tuning for both base and instruct models. Finally, we describe the merging framework that combines task-specific and base models to enhance generalizability.

\subsubsection{\textbf{Instruction Dataset for Instruct Models}}

We adopt the methodology outlined in \cite{wang2023instructuie} to create detailed instructions that aid the model in understanding various tasks. Each task instance in our instruction dataset is structured with three main components: task instruction, input text, and output.

The task instruction provides a guide for the task to be performed based on the input text, along with the expected labels for each task. A comprehensive list of task instructions for each task is provided in Table \ref{table:7} in Appendix \ref{subsec:appendix:a}. We then create each sample from the original dataset by combining the instruction, input, and output in a specific format.

For the Instruct models, we generate a prompt for each model by including the special tokens specified in Table \ref{table:5} in Appendix \ref{subsec:appendix:a}. For test datasets, we enclose the instructions and input within the corresponding special tokens for each model, while using label delimiters like "label:" as suggested in \cite{liu2024tuning}.

\subsubsection{\textbf{Instruction Dataset for Base Models}}
Following the approach of \cite{taori2023stanford}, we designed an instruction scheme to aid the model in understanding the task. As detailed in Table \ref{table:6} in Appendix \ref{subsec:appendix:a}, the instruction scheme includes a generic instruction, an instruction field to guide task completion, and task-specific instructions as shown in Table \ref{table:7} in Appendix \ref{subsec:appendix:a}. The scheme also features an input field and a response field that the LLMs are required to complete.

\subsubsection{\textbf{Instruction Fine-tuning}}
We conducted instruction fine-tuning on all three models using HuggingFace Transformers with 2 epochs and the Adam-w torch fused optimizer. The batch size was set to 2, with an initial learning rate of 2e-4 and warm-up steps comprising 3\% of all training steps. The maximum length of input texts was 1024 for the Phi-3 model and 2048 for the Mistral-7B and Llama3-8B models. All models were fine-tuned on 8 A100 40GB GPUs.

To optimize the fine-tuning process and reduce the computational cost, we employed techniques such as Low-Rank Adaptation (LoRA) with a rank set to 32 and Lora alpha set to 46, alongside quantization \cite{hu2021lora}. LoRA enables fine-tuning the low-rank decomposed factors of the original weight matrices rather than the full matrices, significantly reducing trainable parameters. This approach allows training on less powerful hardware and shortens the total training time \cite{li2023large}.

We experimented with two fine-tuning settings:
\begin{enumerate}
    \item \textbf{Single-Task Fine-Tuning}: \label{section:3.4.3} Each model underwent separate fine-tuning on the dataset specific to each task, as depicted in Figure \ref{fig:flowchart}a, utilizing the specified training settings. These fine-tuned models are subsequently employed in the next section (Model Merging) to enhance the generalizability of the models on the unseen tasks.
    \item \textbf{Multi-Task Fine-Tuning}: Each base and instruct model underwent fine-tuning using a combination of instruction datasets, covering all tasks, as depicted in Figure \ref{fig:flowchart}b. The total dataset consists of 13,194 training examples, as outlined in Table \ref{table:4} in Appendix \ref{subsec:appendix:a}.
\end{enumerate}
   
\subsubsection{\textbf{Model Merging}}
Further fine-tuning of pre-trained models can lead to catastrophic forgetting, degrading general capabilities and reducing performance across tasks \cite{zhang2023instruction, zhang2023multi}.
To mitigate this, leveraging existing pre-trained checkpoints is crucial. Model merging, combining parameters from multiple models trained on specific tasks into a unified model, has become essential. This strategy supports multi-task and continual learning, reducing catastrophic forgetting without retraining costs \cite{goddard2024arcee, yadav2023resolving}.

Our study utilized MergeKit, a library for model merging, employing Task Arithmetic \cite{ilharco2022editing}. This technique involves arithmetic on task vectors, representing differences between fine-tuned models and a common base model. Task Arithmetic enhances generalizability and performance across diverse tasks, effectively mitigating catastrophic forgetting \cite{goddard2024arcee}. We utilize single-task fine-tuned models on LlaMA3-8B and Mistral-7B models, as described in the previous section, and merge them with the vanilla instruct models for the corresponding models (LlaMA3-8B and Mistral-7B) with equal 25\% weight for each of the models \footnote{we were unable to perform model merging for the Phi-3-Instruct model because the Language Model head was updated during fine-tuning and did not match the vanilla model.}, as shown in Figure \ref{fig:flowchart}c. 

\subsubsection{\textbf{Baseline Models}}
We compare the fine-tuned and merged models against three baseline vanilla instruct models and three specialized financial models:
\begin{enumerate}
    \item FinMA-7B \cite{xie2024pixiu}: A 7B parameter model instruction-fine-tuned for financial tasks, including multiple NLP and forecasting tasks.
    \item AdaptLLM-7B \cite{cheng2023adapting}: A model continued pre-trained and fine-tuned on financial news. 
    \item GPT-4 \footnote{We used the GPT-4 model from checkpoint preview of gpt-4-1106-preview and following website \url{https://platform.openai.com/docs/models}}: A robust model from OpenAI. 
\end{enumerate}

In the next section, we will review the results of our experiments in detail, focusing on the comparative performance of the models under different settings, including zero-shot, few-shot, and multi-task fine-tuning scenarios. We will also analyze how these models handle unseen tasks and assess the impact of model merging techniques on improving their generalization capabilities. Our analysis will highlight key trends, identify strengths and weaknesses of each approach, and offer insights into the effectiveness of fine-tuning strategies for enhancing performance on domain-specific financial tasks.

% \input{table_icl}
% table_icl.tex
% table with acc, f1 and sample data
% \begin{table}[h]
%     \centering
%     \caption{In context learning}
%     \label{tab:icl_performance}
%     \begin{tabular}{@{}ll|ccccccc@{}}
%         \toprule
%         \multirow{2}{*}{\textbf{Model}} & \multirow{2}{*}{\textbf{\# example}} & \multicolumn{2}{c}{\textbf{FPB}} & \multicolumn{2}{c}{\textbf{FiQA-SA}} & \textbf{Headline-Dir} & \textbf{FinRED} & \textbf{FOMC} \\ 
%         \cmidrule(r){3-4} \cmidrule(r){5-6} \cmidrule(r){7-7} \cmidrule(r){8-8} \cmidrule(r){9-9}
%         & & \textbf{Acc} & \textbf{F1} & \textbf{Acc} & \textbf{F1} &  \textbf{F1}  & \textbf{F1} & \textbf{F1} \\ 
%         \midrule
%         \multirow{4}{*}{\textit{Llama3-8B}} 
%         & One-shot & 75.5 & 75.1 & 72.5 & 74.8 & 89.1 & \textcolor{red}{4} & 52.8  \\ 
%         & Five-shot & 72.7 & 73.4 & 73.3 & 74.4 & 88.7 & - & 53.2 \\ 
%         & ten-shot & 68.9 & 70.5 & 66.1 & 74.7 & 88.1 & - & 53.6  \\ 
%         \midrule
%         \multirow{3}{*}{\textit{Mistral-7B}} 
%         & One-shot & 74.8 & 72.7 & 48.4 & 57.3 & 86.5 & - & 59.6  \\ 
%         & Five-shot & 74.9 & 71.4 & 47.9 & 54.1 & 85.6 & - & 54.1  \\ 
%         & ten-shot & 68.5 & 69.9 & 37.1 & 53.4 & 84.5 & - & 49.9  \\ 
%         \midrule
%         \multirow{3}{*}{\textit{Phi-3-mini}} 
%         & One-shot & 58.8 & 58.6 & 60.4 & 59.5 & 88.6 & - & 52.9   \\ 
%         & Five-shot & 49.8 & 47.7 & 53.6 & 47. & 87.3 & - & 46.1  \\ 
%         & ten-shot & 26.5 & 51.2 & 51.3 & 49.5 & 81.1 & - & 39.3  \\ 
%         \bottomrule
%     \end{tabular}
% \end{table}

\begin{table}[h]
    \centering
    \begin{tabular}{@{}ll|ccccccc@{}}
        \toprule
        \multirow{2}{*}{\textbf{Model}} & \multirow{2}{*}{\textbf{\# example}} & \multicolumn{2}{c}{\textbf{FPB}} & \multicolumn{2}{c}{\textbf{FiQA-SA}} & \textbf{Headline-Dir}  & \textbf{FOMC} \\ 
        \cmidrule(r){3-4} \cmidrule(r){5-6} \cmidrule(r){7-7} \cmidrule(r){8-8} 
        & & \textbf{Acc} & \textbf{F1} & \textbf{Acc} & \textbf{F1} &  \textbf{F1}   & \textbf{F1} \\ 
        \midrule
        \multirow{4}{*}{\textit{Llama3-8B}} 
        & One-shot & 75.5 & 75.1 & 72.5 & 74.8 & 89.1 & 52.8  \\ 
        & Five-shot & 72.7 & 73.4 & 73.3 & 74.4 & 88.7 &  53.2 \\ 
        & ten-shot & 68.9 & 70.5 & 66.1 & 74.7 & 88.1 &  53.6  \\ 
        \midrule
        \multirow{3}{*}{\textit{Mistral-7B}} 
        & One-shot & 74.8 & 72.7 & 48.4 & 57.3 & 86.5 &  59.6  \\ 
        & Five-shot & 74.9 & 71.4 & 47.9 & 54.1 & 85.6 &  54.1  \\ 
        & ten-shot & 68.5 & 69.9 & 37.1 & 53.4 & 84.5 & 49.9  \\ 
        \midrule
        \multirow{3}{*}{\textit{Phi-3-mini}} 
        & One-shot & 58.8 & 58.6 & 60.4 & 59.5 & 88.6 &  52.9   \\ 
        & Five-shot & 49.8 & 47.7 & 53.6 & 47. & 87.3 &  46.1  \\ 
        & ten-shot & 26.5 & 51.2 & 51.3 & 49.5 & 81.1 &  39.3  \\ 
        \bottomrule
    \end{tabular}
    \caption{Few-shot experiment results on the FBP, FiQA-SA, Headline-Dir, and FOMC datasets using the Llama3-8B, Mistral-7B, and Phi-3-mini instruct models. Results for the FinRED dataset are excluded due to its large label space, which resulted in very low performance (below 5\%).}
    \label{table:2}
\end{table}

% \begin{figure}[ht] % htbp
%     % ICL accuracy diagrams
%     \centering
    
%     \begin{subfigure}[t]{0.49\textwidth}
%         \centering
%         \input{fpb_accuracy_diagram}
%         \caption{In context learning accuracy for FPB dataset}
%     \end{subfigure}
%     \hfill
%     \begin{subfigure}[t]{0.49\textwidth}
%         \centering
%         \input{fiqasa_accuracy_diagram}
%         \caption{In context learning accuracy for FiQA\_SA dataset}
%     \end{subfigure}
    
%     \vspace{1em}
    
%     \begin{subfigure}[t]{0.49\textwidth}
%         \centering
%         \input{headline_accuracy_diagram}
%         \caption{In context learning accuracy for Headline dataset}
%     \end{subfigure}
%     \hfill
%     \begin{subfigure}[t]{0.49\textwidth}
%         \centering
%         \input{fomc_accuracy_diagram}
%         \caption{In context learning accuracy for FOMC dataset}
%     \end{subfigure}
    
%     \caption{Few-shot experiment results (Accuracy) on the FBP, FiQA-SA, Headline-Dir, and FOMC datasets using the Llama3-8B, Mistral-7B, and Phi-3-mini instruct models. Results for the FinRED dataset are excluded due to its large label space, which resulted in a low performance (below 5\%).}
%     \label{fig:2}
% \end{figure}

\begin{figure}[htp]
    \centering
    \includegraphics[width=16cm]{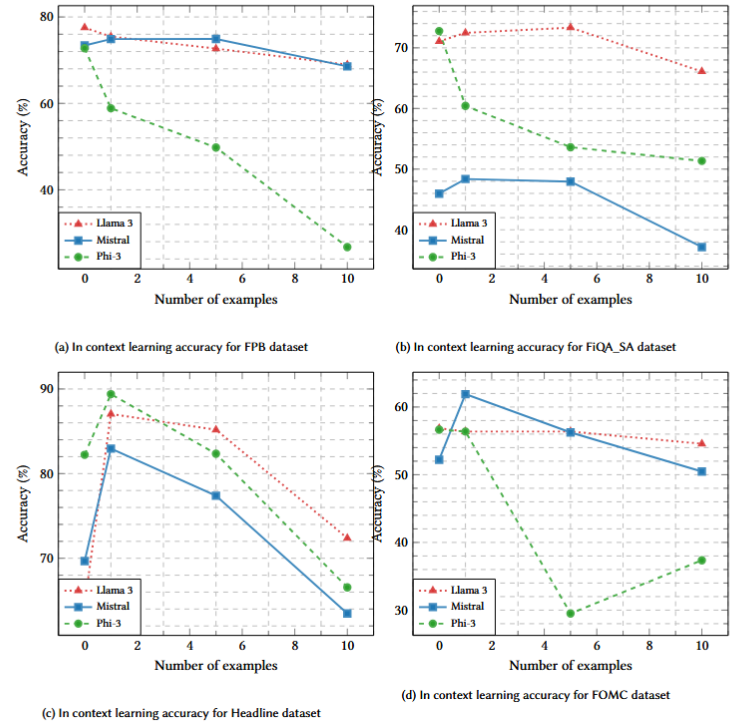}
    \caption{Few-shot experiment results (Accuracy) on the FBP, FiQA-SA, Headline-Dir, and FOMC datasets using the Llama3-8B, Mistral-7B, and Phi-3-mini instruct models. Results for the FinRED dataset are excluded due to its large label space, which resulted in a low performance (below 5\%).}
    \label{fig:3}
\end{figure}

\begin{figure}[htp]
    \centering
    \includegraphics[width=16cm]{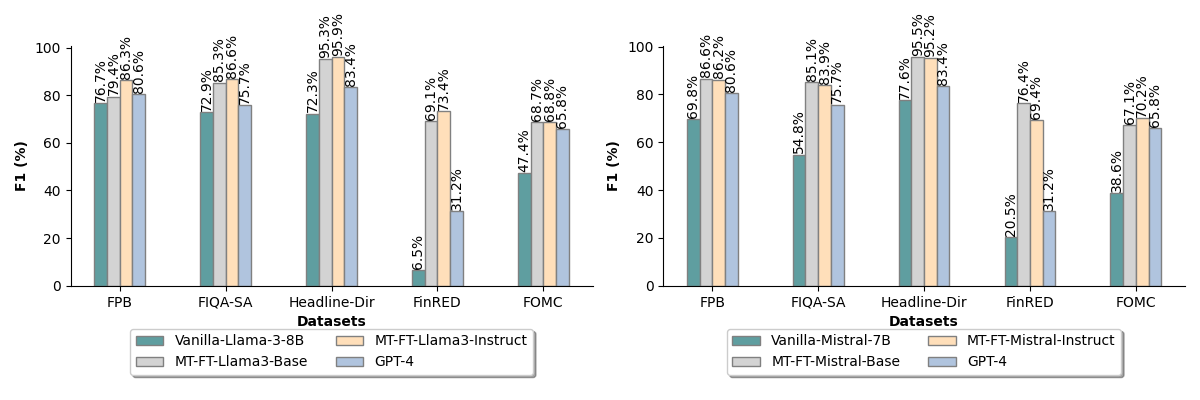}
    \caption{Performance comparison of vanilla models (zero-shot instruct models), multi-task fine-tuned instruct models, multi-task fine-tuned base models, and GPT-4 models on five financial classification datasets for Llama3-8B and Mistral-7B models. F1 score is reported.}
    \label{fig:3}
\end{figure}

\begin{figure}[htp]
    \centering
    \includegraphics[width=12cm]{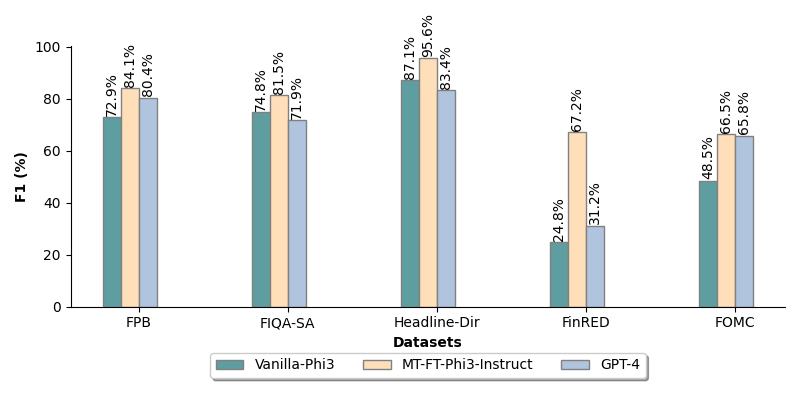}
    \caption{Performance comparison of vanilla models (zero-shot instruct models), multi-task fine-tuned instruct models, and GPT-4 models on five financial classification datasets for Phi-3 model. F1 score is reported.}
    \label{fig:4}
\end{figure}

\section{Results and Analysis}

In this section, we present a comprehensive analysis of the model performance across various experimental setups, including zero-shot, few-shot, multi-task fine-tuning, and model merging techniques. We also provide a detailed error analysis and summarize the key findings.

\subsubsection{\textbf{Evaluation Framework}}

Following standard practices, We evaluate the models using standard classification metrics such as accuracy, and weighted F1 score to assess their performance on financial text classification tasks. The tasks include sentiment analysis, relation extraction, news headline classification, and hawkish-dovish classification. All results are compared against baseline models, including state-of-the-art models like GPT-4 and FinMA-7B, to provide a benchmark for evaluating improvements. We also report and discuss the results of based-fine-tuned, instruct-fine-tuned, and merged models compared to the baselines across four datasets.\\

% For other datasets, we report the weighted F1 score.

% \input{main_table_results_big} 

% table with acc, f1 and sample data
\begin{table}[ht]
    \centering
    \begin{tabular}{@{}ll|ccccccc:ccc@{}}
        \toprule
        \multirow{2}{*}{\textbf{Experiment}} & \multirow{2}{*}{\textbf{Model}} & \multicolumn{2}{c}{\textbf{FPB}} & \multicolumn{2}{c}{\textbf{FiQA-SA}} & \textbf{Headline-Dir} & \textbf{FinRED} & \textbf{FOMC} & \textbf{FinArg} & \textbf{M\&A} & \textbf{SC} \\ 
        \cmidrule(r){3-4} \cmidrule(r){5-6} \cmidrule(r){7-7} \cmidrule(r){8-8} \cmidrule(r){9-9} \cmidrule(r){10-10} \cmidrule(r){11-11} \cmidrule(r){12-12}
        & & \textbf{Acc} & \textbf{F1} & \textbf{Acc} & \textbf{F1} &  \textbf{F1}  & \textbf{F1} & \textbf{F1} & \textbf{F1} &  \textbf{F1} & \textbf{F1} \\ 
        \midrule
        \multirow{4}{*}{\textit{Baseline Models}} 
        & BloombergGPT & - & 51.2 & - & 75.2 & - & - & - & - & - & - \\ 
        & AdaptLLM-7B & - & 62.5 & - & 72.1 & - & - & - & - & - & - \\ 
        & FinMA-7B & 86.0 & 86.1 & 78.3 & 79.2 & - & - & 49.0 & 27.5 & 45.3 & - \\ 
        & GPT-4 & 80.4 & 80.6 & 71.9 & 75.7 & 83.4 & 31.2 & 65.8 & - & - & -  \\ 
        \midrule
        \multirow{3}{*}{\textit{Vanilla Models}} 
        & Llama3-8B & 77.5 & 76.7 & 71.1 & 72.9 & 72.3 & 6.5 & 47.4 & 50.2 & 85.9 & 66.7 \\ 
        & Mistral-7B & 73.4 & 69.8 & 45.9 & 54.8 & 77.6 & 20.5 & 38.6 & 42.2 & 83.6 & 68.8  \\ 
        & Phi-3-mini & 72.8 & 72.9 & 72.7 & 74.8 & 87.1 & 24.8 & 48.5 & 54.1 & 80.1 & 66.5 \\ 
        % \midrule
        % \multirow{2}{*}{\textit{ST-Base-FT}}
        % & Llama3-8B & 84.7 & 84.8 & 88.1 & 88.2 & 95.6 & 74.3 & 67.7 & - & - & - \\ 
        % & Mistral-7B & 85.2 & 85.3 & 85.9 & 86.5 & \textcolor{red}{?} & \textcolor{red}{?} & 71.1 & - & - & - \\ 
        % % & Phi-3-mini & 83.4 & 82.5 & 76.6 & 79.1 & 95.6 & 69.2 & 66.3 & - & - & -  \\ 
        \midrule
        \multirow{2}{*}{\textit{MT-Base-FT}}
        & Llama3-8B & 79.1 & 79.4 & 87.2 & 85.3 & 95.3 & 69.1 & 68.7 & 23.4 & 72.5 & 52.6 \\ 
        & Mistral-7B & 86.8 & 86.6 & 85.5 & 85.1 & 95.5 & 76.4 & 67.1 & 13.4 & 70.9 & 31.8 \\ 
        % & Phi-3-mini & 83.4 & 82.5 & 76.6 & 79.1 & 95.6 & 69.2 & 66.3 & - & - & -  \\ 
        % \midrule
        % \multirow{3}{*}{\textit{ST-Instruct-FT}}
        % & Llama3-8B & 81.1 & 79.5 & 78.7 & 81.4 & 95.5 & 73.2 & 67.6 & - & - & -   \\ 
        % & Mistral-7B & 84.6 & 84.8 & 85.5 & 83.9 & 95.4 & 67.3 & 68.3 & - & - & -  \\ 
        % & Phi-3-mini & 83.4 & 82.5 & 76.6 & 79.1 & 95.6 & 69.2 & 66.3 & - & - & -  \\ 
        \midrule
        \multirow{3}{*}{\textit{MT-Instruct-FT}} 
        & Llama3-8B & 86.2 & 86.3 & 86.4 & 86.6 & 95.9 & 73.4 & 68.4 & 31.1 & 72.9 & 39.2    \\ 
        & Mistral-7B & 86.3 & 86.2 & 85.1 & 83.9 & 95.2 & 69.4 & 70.2 & 35.1 & 76.1 & 57.6   \\ 
        & Phi-3-mini & 84.7 & 84.1 & 79.5 & 81.5 & 95.6 & 67.2 & 66.5 & 53.5 & 77.7 & 64.9 \\
        \midrule
        % \multirow{3}{*}{\textit{Merged Models}} 
        \multirow{2}{*}{\textit{Merged Models}} 
        & Llama3-8B & 80.6 & 80.7 & 75.3 & 78.3 & 93.2 & 44.6 & 60.9 & 54.6 & 74.3 & 65.5  \\ 
        & Mistral-7B & 80.4 & 80.3 & 65.1 & 72.1 & 91.6 & 41.2 & 38.9 & 50.4 & 83.5 & 59.6  \\ 
        % & Phi-3-mini & 40.7 & 87.7 & 24.6 & 84.0 & 16.4 & 81.5 & 14.5 & 70.5 & 18.9 & 30.0  \\
        \bottomrule
    \end{tabular}
     \caption{Main experimental results for four financial classification tasks and three unseen financial classification tasks (FinArg, M\&A, Casual-SC). Vanilla models indicate the zero-shot performance of instruct models. MT-Base-FT refers to multi-task fine-tuned base models, and MT-Instruct-FT refers to multi-task fine-tuned instruct models. Baseline models' results: BloombergGPT results are taken from \cite{wu2023bloomberggpt}, AdaptLLM-7B results are obtained from \cite{cheng2023adapting}, and FinMA-7B results are taken from \cite{xie2024pixiu}. The best results are in \textbf{bold}. }
    \label{table:3}
\end{table}

\subsection{Zero-Shot Performance Analysis}

As shown in Figure \ref{fig:3} and Table \ref{table:3}, the zero-shot performance of the vanilla instruct models (Phi-3, Mistral, and Llama3) was relatively strong on the Financial PhraseBank (FPB), FiQA Sentiment Analysis (FiQA-SA), and headline classification datasets. The Llama3 and Phi-3 models demonstrated notable accuracy in these tasks, while the Mistral model performed below 50\% on datasets that included tweets, likely due to its pre-training corpus lacking similar data.

However, all models struggled on the FOMC and FinRed datasets, showing suboptimal results. The FOMC dataset's complex monetary policy texts and the FinRed dataset's extensive label set of 29 categories posed significant challenges, even for the larger Llama3 model. This performance issue likely stems from the long-tail distributions and the intricate nature of relation extraction tasks that are not well-represented in the models' instruction-tuned training. These findings align with prior observations that such datasets require specialized fine-tuning for optimal performance \cite{wang2023chatgpt, wang2023instructuie}.

On both the FOMC and FinRed datasets, all models exhibited subpar performance. We attribute the poor performance on the FOMC dataset, which contains monetary policy texts, to the texts being more long-tailed compared to other datasets in the pre-training phase, thereby making the downstream task more challenging. This observation aligns with findings from \cite{wang2023chatgpt}. The FinRed dataset, characterized by its large label space (29 labels) and intricate relation extraction tasks, presented substantial challenges. These difficulties likely stem from these tasks not being adequately covered in the instruction-tuned training, which is consistent with the findings of \cite{wang2023instructuie}. \\

Notably, even the Llama3 model, despite its superior architecture and high-quality training data, performed below 10\% on this dataset. This might be due to its strict adherence to instruction-tuned behavior, focusing solely on generating labels without the added contextual explanations that other models like Mistral-7B and Phi-3 provided alongside the labels, possibly due to their Chain-of-Thought (CoT) post-training.\\

\subsection{Few-Shot Results}

To explore the impact of few-shot learning, we conducted experiments using 1-shot, 5-shot, and 10-shot setups, employing three different random seeds for sampling demonstration examples, following the methodology in \cite{wang2023chatgpt} and reported the average performance. The few-shot results, compared against zero-shot results, are summarized in Figure \ref{fig:2} and Table \ref{table:2}.\\

We observed that 1-shot prompting generally improved model performance across most datasets, except for the Phi-3-mini model, where performance slightly decreased on sentiment analysis tasks due to the adverse effects of unrelated examples on smaller language models. Increasing the number of shots to 5 and 10 produced mixed results: while some models like Llama3-8B and Mistral-7B showed only slight declines, others, particularly Phi-3-mini, experienced high variability in performance, indicating its sensitivity to the number of examples used.\\

Notably, all models demonstrated a significant drop in performance on the headline classification dataset when the number of demonstrations increased, even falling below the performance in the zeros-shot setting. This decline suggests that few-shot settings can sometimes introduce noise, leading to a decrease in model accuracy, particularly for smaller models \cite{zhang2023sentiment}.\\

However, models like Llama3-8B and Mistral-7B displayed only slight decreases on other datasets. Among all models, Phi-3-mini showed the highest performance variability when the number of demonstrations increased, underscoring the sensitivity of smaller models to few-shot settings, as previously highlighted in the literature \cite{zhang2023sentiment}. Due to consistently low performance on the FinRed dataset, which did not exceed 10\%, we excluded these results from our analysis, aligning with findings that suggest an increase in input complexity can adversely affect model accuracy on tasks with large label sets\cite{zhang2023sentiment}.

\subsection{Multi-Task Fine-Tuning}
Our multi-task fine-tuning experiments demonstrated more stable behavior across models, as illustrated in Table \ref{table:3} and Figures \ref{fig:3} and \ref{fig:4}. Fine-tuning led to a substantial increase in model performance compared to zero-shot settings in both base and instruct models. Notably, the instruct fine-tuned Llama3-8B model showed a dramatic improvement from 6.5\% to 73.4\% on the relation extraction task (FinRED dataset), with similar gains observed in the Hawkish-Dovish classification tasks (FOMC dataset). Mistral-7B and Phi-3 models showed an average 40\% performance improvement after fine-tuning on both base and instruct models. \\

Additionally, all models outperformed GPT-4 on these finance-related tasks, highlighting the limitations of GPT-4 in specialized domains such as finance and the necessity for task-specific fine-tuning. In sentiment analysis tasks (FPB and FiQA-SA datasets), our models performed comparably to the state-of-the-art FinMA-7B model, surpassing other models like AdalpLLM-7B and BloombergGPT.\\

The results indicated that multi-task fine-tuned instruct models outperformed their base counterparts across all datasets, except the Mistral-7B base model, which slightly outperformed the instruct variant on most tasks. This variance could be attributed to differences in the quality of supervised fine-tuning prompts, suggesting that starting with a robust instruction-tuned checkpoint may offer advantages for domain-specific fine-tuning.\\

The results of the multi-task fine-tuned Phi-3 models also demonstrate performance on par with the other two models, despite having only 3.8 billion parameters. This highlights the potential of the Phi-3 model for fine-tuning downstream financial tasks.
 
\subsection{Generalization Multi-Task Fine-Tuned model to Unseen Tasks}

The almost similar behavior observed in the multi-task fine-tuning of base and instruct models (using the same amount of data without incorporating generic instruction data) on seen tasks indicates that both models benefit nearly equally from fine-tuning on tasks present in the training corpus. However, our results on three unseen financial tasks, as shown in Table \ref{table:3} and Figures \ref{fig:5} and \ref{fig:6}, indicate that multi-task fine-tuned base models exhibit significantly greater performance declines compared to their instruct-tuned counterparts when generic instruction data is absent during fine-tuning. This highlights the risk of performance degradation for base models on tasks they haven't encountered, a trend consistent with prior research on catastrophic forgetting \cite{zhang2023multi, wang2023huatuo}. In contrast, instruct-tuned models retain stronger generalization capabilities, making them more robust in handling unseen tasks.\\

For example, on the Argument Unit Classification task, the Llama3-8B model’s performance drops to 13.4\%, while on the Causal Classification task, the Mistral-7B model experiences a decline to 31.8\%. Notably, the Phi-3-mini model, despite its smaller size of 3.8 billion parameters, shows only a 2\% performance drop, suggesting that it retains its general capabilities better than larger models after fine-tuning. These findings emphasize the effectiveness of instruct-tuned models in maintaining performance across unfamiliar tasks.\\

Overall, while multi-task fine-tuning significantly improves performance on tasks within the training data, it can also increase the risk of overfitting and catastrophic forgetting, particularly when generic instruction data is not included in the training process. These results underscore the importance of diverse instruction data in enhancing a model’s ability to generalize effectively across unseen tasks.

\begin{figure}[htp]
    \centering
    \includegraphics[width=16cm]{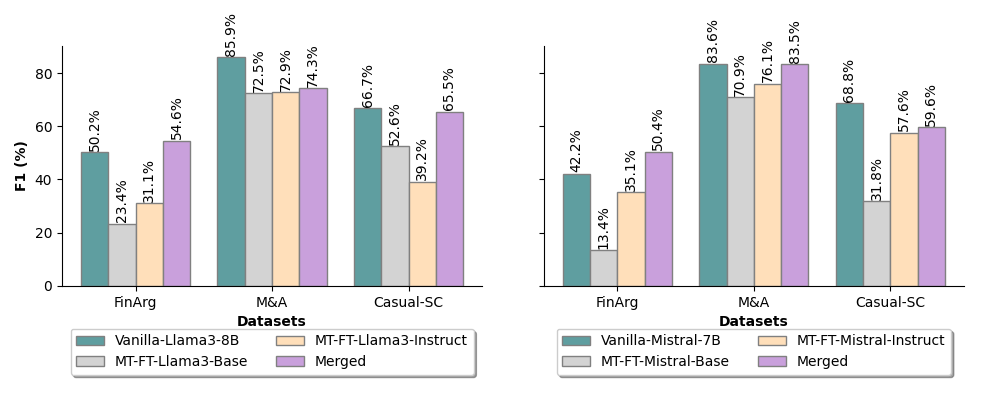}
    \caption{Performance comparison of vanilla models (zero-shot instruct models), multi-task fine-tuned instruct models, multi-task fine-tuned base models, and merged models on three unseen datasets for Llama3-8B and Mistral-7B models. F1 score is reported.}
    \label{fig:5}
\end{figure}

\begin{figure}[htp]
    \centering
    \includegraphics[width=10cm]{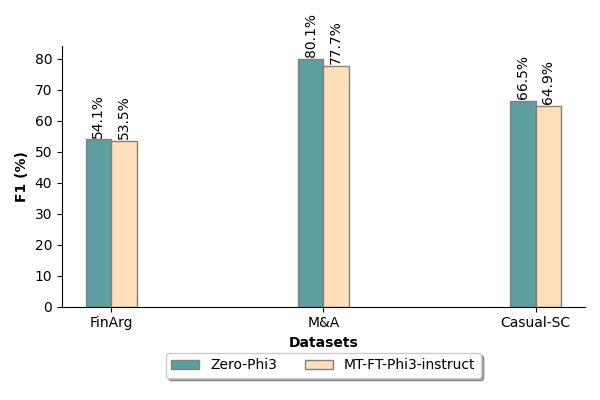}
    \caption{Performance comparison of vanilla models (zero-shot instruct models) and multi-task fine-tuned instruct models on three unseen datasets for the Phi-3 model. F1 score is reported.}
    \label{fig:6}
\end{figure}

\subsection{Merged Models}

To mitigate the degradation of zero-shot performance on unseen tasks, we utilized the MergeKit framework (arithmetic method), which combines single-task fine-tuned models with the vanilla instruct model. The merging approach significantly enhances performance on unseen tasks, with results often surpassing the original zero-shot performance, significant for the FinArg dataset. 

The results for the Llama3-8B and Mistral-7B models are shown in Figure \ref{fig:5}. For example, the merged models outperformed both the fine-tuned, base models and zero-shot in handling complex tasks, suggesting that the integration of diverse task knowledge leads to a more generalized and robust performance. The performance of single-task fine-tuning on both base and instruct models is presented in Appendix \ref{subsec:appendix:b}. \\

The merging strategy helps to preserve the benefits of fine-tuning while reducing the risk of catastrophic forgetting on unseen datasets without incorporating a generic instruction dataset, which increases computational cost and time.

While fine-tuning improves performance over base models, merging fine-tuned models further enhances their ability to handle unseen tasks by a) Combining diverse knowledge sources, b) Reducing overfitting, and c) Leveraging complementary strengths of different models. The regularization effect from merging also helps to produce a more generalized and robust model performance across diverse tasks.

We can also observe that the performance drop of the Phi3 base model compared to its fine-tuned counterpart is not much compared to the other two models (Llama3 and Mistral), as indicated in \ref{fig:6}. This could be due to the following reasons. 
\begin{itemize}
    \item Model Architecture: Phi3 may have an architecture that is more resistant to overfitting during fine-tuning. It's possible that Phi3 has better built-in regularization mechanisms or a structure that promotes better generalization.
    \item Pre-training Approach: The pre-training approach used for Phi3 might result in more robust general knowledge that is less easily overwritten during fine-tuning. This could lead to better retention of broad capabilities even after task-specific training.
    \item Model Size: If Phi3 is a smaller model compared to Mistral and Llama3, it might have less capacity to overfit to the fine-tuning data, paradoxically leading to better generalization on unseen tasks.
\end{itemize}

The significant performance drops in Mistral and Llama3 suggest that these models might be more susceptible to overfitting during fine-tuning, possibly due to larger model sizes, different architectures, or fine-tuning approaches that allow for more dramatic changes to the model's parameters.
This difference highlights the importance of careful fine-tuning practices and the need to consider the trade-offs between task-specific performance and general capabilities when adapting large language models to specific tasks. It also demonstrates why techniques like model merging can be valuable in recovering and combining the strengths of different model versions.

\subsection{Error Analysis and Model Insights}

Our error analysis reveals significant improvements in fine-tuned models across various tasks, particularly in financial sentiment detection, relation extraction, news headline classification, and hawkish-dovish classification. These enhancements stem from the models' increased ability to capture domain-specific nuances, understand hierarchical corporate structures, and interpret economic indicators. In financial sentiment detection, fine-tuned models excel at identifying subtle cues and handling mixed sentiments more effectively. For relation extraction and news headline classification, fine-tuning leads to a better interpretation of corporate ties and financial movements. Similarly, in the hawkish-dovish classification, the models display a stronger grasp of monetary policies and their economic implications. Below, we provide detailed case studies involving Phi-3, Mistral, and Llama3 models, highlighting performance improvements across these key areas.

%In this section, we analyze the effect of fine-tuning across various datasets and models. We provide case study examples to demonstrate how fine-tuning improves the performance of Phi-3, Mistral, and Llama3 models. These improvements are organized into key categories.\\

%Through detailed analysis, we identified that fine-tuned models show notable improvements in several areas, including:

% Financial Sentiment Detection: Fine-tuned models demonstrated a stronger grasp of subtle financial cues, handling complex and mixed sentiments more effectively.
%Relation Extraction: Improved understanding of hierarchical relationships and domain-specific nuances enabled more accurate classification of corporate structures and financial ties.
% Monetary Policy Interpretation: Models displayed enhanced capabilities in recognizing economic indicators and contextualizing monetary policies in FOMC statements.

\subsubsection{Financial Sentiment Classification Result Analysis}

Fine-tuning yielded significant performance gains in financial sentiment classification, particularly in the following areas:

\paragraph{1. Domain-Specific Knowledge}

Fine-tuned models demonstrated improved comprehension of financial language and context. For example, in Table \ref{tab:Domain_Specific_Knowledge_FPB}, fined-tuned models correctly interpreted growth statements in a financial context, where the base models failed to capture the nuance.

\begin{table}[ht]
    \centering
    \begin{tabular}{|p{12cm}|}
        \hline
        \texttt{\textbf{Sentence:} "Finnish Bore that is owned by the Rettig family has grown recently through the acquisition of smaller shipping companies."} \\
        \hline
        \texttt{\textbf{Ground Truth Label:} Positive \newline \textbf{Model Predictions:} \newline \textcolor{mygreen}{Phi-3 Base: Neutral}, \textcolor{blue}{Phi-3 Fine-tuned: Positive} \newline \textcolor{mygreen}{Mistral Base: Neutral}, \textcolor{blue}{Mistral Fine-tuned: Positive}\newline
        \textcolor{mygreen}{Llama3 Base: Neutral}, \textcolor{blue}{Llama3 Fine-tuned: Positive}} \\
        \hline
    \end{tabular}
    \caption{Domain-Specific Knowledge Example - FPB dataset}
    \label{tab:Domain_Specific_Knowledge_FPB}
\end{table}

\paragraph{2. Improved Sentiment Detection}
Fine-tuned models showed enhanced capability in detecting subtle sentiment cues within financial statements, such as recognizing negative sentiment related to financial losses, as illustrated in Table \ref{tab:Improved_Sentiment_Detection_FPB}.

\begin{table}[ht]
    \centering
    \begin{tabular}{|p{12cm}|}
        \hline
        \texttt{\textbf{Sentence:} "Finnair said that the cancellation of flights would cause daily losses of €2.5 million US\$3 million."} \\
        \hline
        \texttt{\textbf{Ground Truth Label:} Negative \newline\textbf{Model Predictions:} \newline \textcolor{mygreen}{Mistral Base: Neutral}, \textcolor{blue}{Mistral Fine-tuned: Negative}\newline 
        \textcolor{mygreen}{Llama3 Base: Neutral}, \textcolor{blue}{Llama3 Fine-tuned: Negative}\newline
        \textcolor{mygreen}{Phi-3 Base: Neutral}, \textcolor{blue}{Phi-3 Fine-tuned: Negative}} \\
        \hline
    \end{tabular}
    \caption{Improved Sentiment Detection Example - FPB dataset}
    \label{tab:Improved_Sentiment_Detection_FPB}
\end{table}

\paragraph{3. Better Handling of Mixed Sentiment}
When encountering statements containing positive and negative elements, fine-tuned models, particularly Mistral, identify the prevailing sentiment better, as shown in Table \ref{tab:Better_Handling_Mixed_Sentiment_FPB}.

\begin{table}[!h]
    \centering
    \begin{tabular}{|p{12cm}|}
        \hline
        \texttt{\textbf{Sentence:} "Kalnapilio-Tauro Grupe, which is owned by Denmark's Royal Unibrew, raised its market share... by 14.5 percent to 40.5 million liters."} \\
        \hline
        \texttt{\textbf{Ground Truth Label:} Positive \newline \textbf{Model Predictions:} \newline \textcolor{mygreen}{Phi-3 Base: Positive}, \textcolor{blue}{Phi-3 Fine-tuned: Positive} \newline \textcolor{mygreen}{Mistral Base: Neutral}, \textcolor{blue}{Mistral Fine-tuned: Positive} \newline \textcolor{mygreen}{Llama3 Base: Positive}, \textcolor{blue}{Llama3 Fine-tuned: Positive}} \\
        \hline
    \end{tabular}
    \caption{Better Handling of Mixed Sentiment Example - FPB dataset}
    \label{tab:Better_Handling_Mixed_Sentiment_FPB}
\end{table}

\paragraph{4. Reduced Tendency to Default to Neutral}
Fine-tuned models show greater confidence in assigning sentiment to complex statements, often correctly identifying positive or negative impacts in financial contexts where base models defaulted to neutral, as demonstrated in Table \ref{tab:Reduced_Tendency_FPB}.

\begin{table}[th]
    \centering
    \begin{tabular}{|p{12cm}|}
        \hline
        \texttt{\textbf{Sentence:} "The transaction will have a positive impact of around EUR2m on earnings, which Ruukki will recognize during the fourth quarter."} \\
        \hline
        \texttt{\textbf{Ground Truth Label:} Positive \newline \textbf{Model Predictions:} \newline \textcolor{mygreen}{Phi-3 Base: Neutral}, \textcolor{blue}{Phi-3 Fine-tuned: Positive} \newline \textcolor{mygreen}{Mistral Base: Neutral}, \textcolor{blue}{Mistral Fine-tuned: Positive}\newline
        \textcolor{mygreen}{Llama3 Base: Neutral}, \textcolor{blue}{Llama3 Fine-tuned: Positive}}
        \\
        \hline
    \end{tabular}
    \caption{Reduced Tendency to Default to Neutral Example - FPB dataset}
    \label{tab:Reduced_Tendency_FPB}
\end{table}

\paragraph{5. Understanding of Financial Metrics}
Fine-tuned models exhibited an enhanced ability to interpret financial metrics and their implications, as evidenced in Table \ref{tab:Understand_Financial_Metrics_FPB}, where models accurately assessed financial performance indicators.

\begin{table}[h]
    \centering
    \begin{tabular}{|p{12cm}|}
        \hline
        \texttt{\textbf{Sentence:} "Cash flow from operations rose to EUR 52.7 mn from EUR 15.6 mn in 2007."} \\
        \hline
        \texttt{\textbf{Ground Truth Label:} Positive \newline \textbf{Model Predictions:} \newline \textcolor{mygreen}{Phi-3 Base: Positive}, \textcolor{blue}{Phi-3 Fine-tuned: Positive} \newline \textcolor{mygreen}{Mistral Base: Neutral}, \textcolor{blue}{Mistral Fine-tuned: Positive} \newline \textcolor{mygreen}{Llama3 Base: Positive}, \textcolor{blue}{Llama3 Fine-tuned: Positive}} \\
        \hline
    \end{tabular}
    \caption{Understanding of Financial Metrics Example - FPB dataset}
    \label{tab:Understand_Financial_Metrics_FPB}
\end{table}

\subsubsection{FOMC Classification Result Analysis}

Fine-tuned models also showed marked improvements in interpreting monetary policy statements from the Federal Open Market Committee (FOMC). This was particularly evident in their ability to interpret economic indicators and broader economic contexts.

\paragraph{1. Understanding Economic Indicators}

Fine-tuned models better interpret the impact of economic indicators and the relationship between them like unemployment rates.

For example, as illustrated in Table \ref{tab:Understanding Economic Indicators Example - FOMC dataset}, the base models did not interpret the economic indicators correctly and classified this example as neutral. In contrast, fine-tuned models correctly identify it as Hawkish, recognizing that low unemployment and robust job gains typically lead to tighter monetary policy to prevent overheating.
% For example, as illustrated in Table \ref{tab
% Economic Indicators Example - FOMC dataset}, fine-tuned models demonstrated a better understanding of the relationship between unemployment rates and monetary policy.  Fine-tuned models correctly classified economic indicators like unemployment as hawkish, while base models misclassified them as neutral.

\begin{table}[h]
    \centering
    \begin{tabular}{|p{12cm}|}
        \hline
        \texttt{\textbf{Sentence:} "The unemployment rate edged down to 3.5 percent, and job gains have been robust in recent months."}\\
        \hline
        \texttt{\textbf{Ground Truth Label:} Hawkish \newline
        \textbf{Model Predictions:} \newline \textcolor{mygreen}{Phi-3 Base: Neutral}, \textcolor{blue}{Phi-3 Fine-tuned: Hawkish} \newline \textcolor{mygreen}{Mistral Base: Neutral}, \textcolor{blue}{Mistral Fine-tuned: Hawkish} \newline \textcolor{mygreen}{Llama3 Base: Neutral}, \textcolor{blue}{Llama3 Fine-tuned: Hawkish}} \\
        \hline
    \end{tabular}
    \caption{Understanding Economic Indicators Example - FOMC dataset}
    \label{tab:Understanding Economic Indicators Example - FOMC dataset}
\end{table}

\paragraph{2. Improved Contextual Interpretation}
Fine-tuned models show marked improvement in interpreting broader economic contexts.

\begin{table}[!h]
    \centering
    \begin{tabular}{|p{12cm}|}
        \hline
        \texttt{\textbf{Sentence:} "Inflation has been running persistently below the Committee's longer-run goal of 2 percent."} \\
        \hline
        \texttt{\textbf{Ground Truth Label:} Dovish \newline
        \textbf{Model Predictions:} \newline \textcolor{mygreen}{Phi-3 Base: Neutral}, \textcolor{blue}{Phi-3 Fine-tuned: Dovish} \newline \textcolor{mygreen}{Mistral Base: Neutral}, \textcolor{blue}{Mistral Fine-tuned: Dovish} \newline \textcolor{mygreen}{Llama3 Base: Neutral}, \textcolor{blue}{Llama3 Fine-tuned: Dovish}} \\
        \hline
    \end{tabular}
    \caption{Improved Contextual Interpretation Example - FOMC dataset}
    \label{tab:Improved Contextual Interpretation Example - FOMC dataset}
\end{table}

The example of table \ref{tab:Improved Contextual Interpretation Example - FOMC dataset}, indicates the base models often interpret the examples incorporating economic contexts as neutral, while fine-tuned models correctly identify it as Dovish, recognizing that ``sustained expansion'' typically implies a continuation of accommodative policy.

\paragraph{3. Reduced Misclassification of Complex Statements}
Fine-tuned models display improvement in handling statements with mixed signals.

\begin{table}[!h]
    \centering
    \begin{tabular}{|p{12cm}|}
        \hline
        \texttt{\textbf{Sentence:} "Although household spending has been rising at a strong pace, business fixed investment and exports remain weak."} \\
        \hline
        \texttt{\textbf{Ground Truth Label:} Dovish \newline
        \textbf{Model Predictions:} \newline \textcolor{mygreen}{Phi-3 Base: Neutral}, \textcolor{blue}{Phi-3 Fine-tuned: Dovish} \newline \textcolor{mygreen}{Mistral Base: Neutral}, \textcolor{blue}{Mistral Fine-tuned: Dovish} \newline \textcolor{mygreen}{Llama3 Base: Hawkish}, \textcolor{blue}{Llama3 Fine-tuned: Dovish}} \\
        \hline
    \end{tabular}
    \caption{Reduced Misclassification of Complex Statements Example - FOMC example}
    \label{tab:Reduced Misclassification of Complex Statements Example - FOMC example}
\end{table}

Table \ref{tab:Reduced Misclassification of Complex Statements Example - FOMC example} illustrates that base models often struggle with such mixed signals in complex statements, frequently defaulting to Neutral. Fine-tuned models are better at weighing these factors, often correctly identifying this as a Dovish statement due to the emphasis on weak areas of the economy.

\paragraph{4. Handling of Edge Cases}
Fine-tuned models show improved performance on edge cases or unusual phrasings that might confuse base models.

\begin{table}[!h]
    \centering
    \begin{tabular}{|p{12cm}|}
        \hline
        \texttt{\textbf{Sentence:} "The Committee will be patient as it determines what future adjustments to the target range for the federal funds rate may be appropriate."} \\
        \hline
        \texttt{\textbf{Ground Truth Label:} Dovish \newline
        \textbf{Model Predictions:} \newline \textcolor{mygreen}{Phi-3 Base: Neutral}, \textcolor{blue}{Phi-3 Fine-tuned: Dovish} \newline \textcolor{mygreen}{Mistral Base: Neutral}, \textcolor{blue}{Mistral Fine-tuned: Dovish} \newline \textcolor{mygreen}{Llama3 Base: Neutral}, \textcolor{blue}{Llama3 Fine-tuned: Dovish}} \\
        \hline
    \end{tabular}
    \caption{Handling of Edge Cases Example - FOMC dataset}
    \label{tab:Handling of Edge Cases Example - FOMC dataset}
\end{table}

Example of table \ref{tab:Handling of Edge Cases Example - FOMC dataset} shows base models often interpret sentences with unusual phrases as Neutral, while fine-tuned models correctly identify it as Dovish, recognizing that ``patience'' in this context often implies a willingness to maintain accommodative policy.

% ////////////////////////////////////////FinnRED///////////////////////////////////////////////////////////////
\subsubsection{FinRED Dataset Error and Improvement Analysis}
In the domain of relation extraction from financial texts, fine-tuned models displayed significant improvements across several dimensions:

\paragraph{1. Domain-Specific Knowledge Acquisition}
Fine-tuned models have been exposed to a large volume of financial and corporate relationship data, allowing them to learn domain-specific nuances that base models may not have encountered. 

\begin{table}[h]
    \centering
    \begin{tabular}{|p{12cm}|}
        \hline
        \texttt{\textbf{Sentence:} "For more than 25 years, Stratasys Ltd. ( SSYS ) has been a defining force and dominant player in 3D printing and additive manufacturing – shaping the way things are made."} \\
        \hline
        \texttt{\textbf{Ground Truth Label:} Product/material produced \newline\textbf{Model Predictions:} \newline \textcolor{mygreen}{Phi-3 Base: Developer}, \textcolor{blue}{Phi-3 Fine-tuned: Product/material produced} \newline \textcolor{mygreen}{Mistral Base: Manufacturer}, \textcolor{blue}{Mistral Fine-tuned: Product/material produced} \newline \textcolor{mygreen}{Llama3 Base: Manufacturer}, \textcolor{blue}{Llama3 Fine-tuned: Product/material produced}} \\
        \hline
    \end{tabular}
    \caption{Domain-Specific Knowledge Acquisition Example - FinRED dataset}
    \label{tab:Domain-Specific Knowledge Acquisition Example - FinRED dataset}
\end{table}

In the example of table \ref{tab:Domain-Specific Knowledge Acquisition Example - FinRED dataset}, in the Stratasys-3D printing relationship, the fine-tuned models correctly identify the "product/material produced" relationship, understanding that Stratasys is a company that produces 3D printing technologies. The base models, lacking this specific industry knowledge, misclassify the relationship as "developer" or "manufacturer", which are less precise in describing the company's role in the 3D printing industry. This demonstrates how fine-tuned models have acquired domain-specific knowledge about the 3D printing industry and the relationships between companies and their core technologies.

\paragraph{2. Contextual Interpretation}
Fine-tuned models develop an enhanced ability to interpret contextual cues within sentences, leading to more accurate relationship classification.

\begin{table}[h]
    \centering
    \begin{tabular}{|p{12cm}|}
        \hline
        \texttt{\textbf{Sentence:} "Miner Glencore surged 9 percent, having dropped 30 percent in the previous session to an all-time low."} \\
        \hline
        \texttt{\textbf{Ground Truth Label:} Industry \newline\textbf{Model Predictions:} \newline \textcolor{mygreen}{Phi-3 Base: None}, \textcolor{blue}{Phi-3 Fine-tuned: Industry} \newline \textcolor{mygreen}{Mistral Base: Industry}, \textcolor{blue}{Mistral Fine-tuned: Industry} \newline \textcolor{mygreen}{Llama3 Base: Owned by}, \textcolor{blue}{Llama3 Fine-tuned: Industry}} \\
        \hline
    \end{tabular}
    \caption{Contextual Interpretation Example - FinRED dataset}
    \label{tab:Contextual Interpretation Example - FinRED dataset}
\end{table}

The example of table \ref{tab:Contextual Interpretation Example - FinRED dataset}, in the Glencore-mining case, the fine-tuned model correctly interprets "mining and trading company" to classify Glencore's relationship to mining as "industry". The base model, possibly confused by the complex sentence structure, and misclassified this as a corporate structure relationship ("parent organization" or "subsidiary").

\paragraph{3. Hierarchical Relationship Understanding}
Through exposure to various corporate structures, fine-tuned models better understand the nuances of organizational hierarchies and roles.

\begin{table}[h]
    \centering
    \begin{tabular}{|p{12cm}|}
        \hline
        \texttt{\textbf{Sentence:} "Gucci owner Kering (Swiss: KER.SW - news) published its financial report for the quarter."} \\
        \hline
        % \texttt{\textbf{Label (Ground Truth):} "parent organization" (describing the relationship between Kering and Gucci).} \\
        % \hline
        \texttt{\textbf{Ground Truth Label:} parent organization \newline
        \textbf{Model Predictions:} \newline 
        \textcolor{mygreen}{Phi-3 Base: owner of}, \textcolor{blue}{Phi-3 Fine-tuned: parent organization} \newline
        \textcolor{mygreen}{Mistral Base: owner of}, \textcolor{blue}{Mistral Fine-tuned: parent organization} \newline
        \textcolor{mygreen}{Llama3 Base: owned by}, \textcolor{blue}{Llama3 Fine-tuned: owned by}} \\
        \hline
    \end{tabular}
    \caption{Hierarchical Relationship Understanding Example - FinRED dataset}
    \label{tab:Hierarchical Relationship Understanding Example - FinRED dataset}
\end{table}

In the case presented in table \ref{tab:Hierarchical Relationship Understanding Example - FinRED dataset}, the fine-tuned models identified the nuanced hierarchical relationship ("parent organization") between Kering and Gucci, while the base models defaulted to "owner," failing to capture the specific corporate structure. This shows the improved ability of fine-tuned models to comprehend the complexities of organizational hierarchies. This examples shows how fine-tuned models develop a deeper understanding of hierarchical relationships, particularly in corporate structures, where roles and organizational layers are not always straightforward.

\paragraph{4. Entity-Relationship Mapping}

Fine-tuned models develop a more sophisticated understanding of how different entities (companies, products, people) typically relate to each other in the corporate world.

\begin{table}[h]
    \centering
    \begin{tabular}{|p{12cm}|}
        \hline
        \texttt{\textbf{Sentence:} "First Eagle is currently owned by members of the founding families."} \\
        \hline
        \texttt{\textbf{Ground Truth Label:} Industry \newline
        \textbf{Model Predictions:} \newline 
        \textcolor{mygreen}{Phi-3 Base: Owner of}, \textcolor{blue}{Phi-3 Fine-tuned: Industry}\newline
        \textcolor{mygreen}{Llama3 Base: Owner of}, \textcolor{blue}{Llama3 Fine-tuned: Industry} \newline
        \textcolor{mygreen}{Mistral Base: Owner of}, \textcolor{blue}{Mistral Fine-tuned: Industry}
        } \\
        \hline
    \end{tabular}
    \caption{Entity-Relationship Mapping Example - FinRED dataset}
    \label{tab:Entity-Relationship Mapping Example - FinRED dataset}
\end{table}

The example of table \ref{tab:Entity-Relationship Mapping Example - FinRED dataset} shows that the fine-tuned models correctly understood the ownership structure of First Eagle, while the base models confused the relationship and wrongly focused on ownership, highlighting the fine-tuned models' better grasp of corporate entity-relationship mapping.\

\paragraph{5. Reduction of Overgeneralization}

Fine-tuned models learn to avoid defaulting to common but incorrect relationships when faced with ambiguity, a problem often seen in base models.

\begin{table}[h]
    \centering
    \begin{tabular}{|p{12cm}|}
        \hline
        \texttt{\textbf{Sentence:} "Saudi Arabian budget carrier flynas, which made a name for itself as a low-cost airline."} \\
        \hline
        \texttt{\textbf{Ground Truth Label:} Product/material produced \newline
        \textbf{Model Predictions:} \newline 
        \textcolor{mygreen}{Phi-3 Base: Manufacturer}, \textcolor{blue}{Phi-3 Fine-tuned: Product/material produced} \newline
        \textcolor{mygreen}{Mistral Base: Manufacturer}, \textcolor{blue}{Mistral Fine-tuned: Product/material produced} \newline
        \textcolor{mygreen}{Llama3 Base: Manufacturer}, \textcolor{blue}{Llama3 Fine-tuned: Product/material produced}} \\
        \hline
    \end{tabular}
    \caption{Reduction of Overgeneralization Example - FinRED dataset}
    \label{tab:Reduction of Overgeneralization Example - FinRED dataset}
\end{table}

In this case showen in table \ref{tab:Reduction of Overgeneralization Example - FinRED dataset}, the relationship between flynas and the products or services it provides, with the ground truth label being "product/material produced." Here, the base models overgeneralized the relationship by assuming that flynas was a "manufacturer," which is a common but incorrect assumption, while the fine-tuned models correctly captured the nuanced relationship of flynas producing a service (low-cost flights).

\paragraph{6. Handling of Complex Sentences}

Financial texts often contain complex, information-dense sentences. Fine-tuned models learn to parse these more effectively.

\begin{table}[h]
    \centering
    \begin{tabular}{|p{12cm}|}
        \hline
        \texttt{\textbf{Sentence:} "Excluding \$4.4 million of costs associated with the strategic restructuring initiative recorded in the six months ended June 30, 2019, our selling, general and administrative expenses increased \$6.4 million primarily due to increased selling and marketing expenses in connection with the commercial launch of ANJESO."} \\
        \hline
        \texttt{\textbf{Ground Truth Label:} Industry\newline
            \textbf{Model Predictions:} \newline \textcolor{mygreen}{Phi-3 Base: Product/material produced}, \textcolor{blue}{Phi-3 Fine-tuned: Industry} \newline \textcolor{mygreen}{Mistral Base: Marketing}, \textcolor{blue}{Mistral Fine-tuned: Industry} \newline \textcolor{mygreen}{Llama3 Base: None}, \textcolor{blue}{Llama3 Fine-tuned: Industry}
        } \\
        \hline
    \end{tabular}
    \caption{Handling of Complex Sentences Example - FinRED dataset}
    \label{tab:Handling of Complex Sentences Example - FinRED dataset}
\end{table}

In the example of table \ref{tab:Handling of Complex Sentences Example - FinRED dataset}, despite the complexity of this financial statement, the fine-tuned models correctly identify the relationship between the company and its industry. They focus on the relevant information about selling and marketing expenses, inferring an industry relationship. The base models either misclassify or fail to identify any relationship in this complex sentence.

Through these case studies, it's evident that fine-tuning significantly improves model performance across various financial and economic tasks. Fine-tuned models exhibit better sentiment detection, relationship extraction, and contextual understanding, leading to more accurate predictions and analyses. However, fine-tuning LLM models can lead to catastrophic forgetting and compromise the model performance on unseen tasks which can be improved by using techniques such as merging models. 

\subsubsection{Merging Models for Improved Performance on unseen tasks-datasets.} Merging models helps improve performance on unseen tasks for several reasons. 

\paragraph{1. Diverse knowledge integration.}
Merged models combine knowledge from multiple models trained on different tasks or datasets, allowing them to leverage a broader range of information and patterns.

\begin{table}[h]
    \centering
    \begin{tabular}{|p{12cm}|}
        \hline
        \texttt{\textbf{Sentence from M\&A dataset:} "Autodesk, a maker of design and architecture software, has reached an agreement to acquire construction technology start-up PlanGrid for USD 875.00 million net of cash."} \\
        \hline
        \texttt{\textbf{Ground Truth Label:} Rumor\newline 
            \textbf{Model Predictions:} \newline \textcolor{mygreen}{Llama3 Base: Complete}, \textcolor{blue}{Llamma3 Fine-tuned: Rumor} \newline \textcolor{mygreen}{Mistral-Llama3 Merged: Complete}
        } \\
        \hline
    \end{tabular}
    \caption{Example of Merging Models to Handle Unseen Tasks from M\&A dataset}
\end{table}

For the above entry, the Llama3 fine-tuned model incorrectly classified it as a rumor, while both the base and merged models correctly identified it as complete. This suggests that the fine-tuned model might have overfit to certain patterns, while the merged model was able to correct this error.

\paragraph{2. Reduced Overfitting and Regularization.}
Merging can act as a form of regularization, helping to average out overspecialized patterns and less likely to overfit to specific patterns in the fine-tuning data. It promotes more general features, making the models more robust on unseen tasks.

\begin{table}[h]
    \centering
    \begin{tabular}{|p{12cm}|}
        \hline
        \texttt{\textbf{Sentence from FinArg dataset:} "It's a global number and we are very glad to have the success of the FBA program."} \\
        \hline
        \texttt{\textbf{Ground Truth Label:} Claim \newline
        \textbf{Model Predictions:} \newline 
        \textcolor{mygreen}{Llama3 Fine-tuned: Neutral}, \textcolor{blue}{Llama3 Merged: Claim} \newline
        \textcolor{mygreen}{Mistral Fine-tuned: Premise}, \textcolor{blue}{Mistral Merged: Claim}} \\
        \hline
    \end{tabular}
    \caption{Reduced Overfitting and Regularization Example - FinArg dataset}
    \label{tab:Reduced overfitting and Regularization Example - FinArg dataset}
\end{table}

As shown in table \ref{tab:Reduced overfitting and Regularization Example - FinArg dataset}, the merged model avoids the overfitting and classify the sentence correctly as ``claim'', where the fine-tuned model misclassifies this example as a ``premise''.

\paragraph{3. Complementary strengths.} Different models may excel at different aspects of the task. Merging allows the combined model to leverage the strengths of each constituent model as shown in table \ref{tab:Complementary strengths Example - SC dataset}
\begin{table}[h]
    \centering
    \begin{tabular}{|p{12cm}|}
        \hline
        \texttt{\textbf{Sentence from SC dataset:} " Seth Golden, , Daily Articles, 0  The first trading day post the Saudi Arabia oil field bombings and speculative production slowdown pushed crude oil future (CLF) prices up roughly 15\%, with equity prices moving lower Monday."} \\
        \hline
        \texttt{\textbf{Ground Truth Label:} Causal \newline
        \textbf{Model Predictions:} \newline 
        \textcolor{mygreen}{Llama3 Fine-tuned: Noise}, \textcolor{blue}{Llama3 Merged: Causal} \newline
        \textcolor{mygreen}{Mistral Fine-tuned: Noise}, \textcolor{blue}{Mistral Merged: Causal}} \\
        \hline
    \end{tabular}
    \caption{Complementary strengths Example - SC dataset}
    \label{tab:Complementary Strengths Example - SC dataset}
\end{table}

\section{CONCLUSION AND FUTURE WORK}

In this study, we evaluated the in-context learning (ICL) capabilities of three small instruct models—Llama3-8B, Mistral-7B, and Phi-3—across various financial classification tasks. The results highlighted variability in model performance with increasing numbers of shots. Overall, ICL did not significantly enhance the models’ ability to learn downstream tasks from examples, particularly for smaller models. Among the three, Llama3-8B exhibited marginally better performance, though the gains were not substantial.

Beyond ICL, our primary focus was on the instruct fine-tuning of both the base and instruct versions of these models for four key financial tasks: sentiment analysis, news headline classification, relation extraction, and hawkish-dovish classification. The results demonstrate that multi-task fine-tuning of instruct models significantly improves task-specific performance, particularly for complex tasks such as relation extraction and hawkish-dovish classification. These findings underscore the potential of fine-tuning smaller LLMs for domain-specific tasks in finance.

Notably, the study also revealed that multi-task fine-tuned base models, such as Mistral-7B and Llama3-8B, exhibited greater performance degradation on unseen tasks compared to their instruct-tuned counterparts. This suggests that starting with instruct models, which have already been fine-tuned on a variety of tasks, provides a more robust foundation for maintaining generalization capabilities. Notably, the Phi-3 model, despite its smaller size (3.8 billion parameters), showed minimal performance decline on unseen tasks, indicating that it remains highly capable after fine-tuning for specialized financial tasks.

Looking ahead, we plan to extend this work by exploring more complex financial tasks, such as question answering, which involves retrieving numerical data from tables, and stock market prediction. These tasks will allow us to further investigate instruct fine-tuning on both small base and instruct models. Additionally, as model merging is gaining traction, we aim to experiment with advanced merging techniques like Dare and Tie to assess their effectiveness in mitigating performance degradation on unseen tasks, potentially leading to more versatile and resilient models.

\bibliographystyle{ACM-Reference-Format}
\bibliography{instruct}
% \appendix
\begin{appendices}

\appendix
\section*{Appendix}

\subsection{Appendix A}
\label{subsec:appendix:a}
We present statistics regarding the datasets utilized for instruction fine-tuning of both base and instruct models. Subsequently, we provide descriptions of the instructions and prompt templates employed for each task during instruction fine-tuning.
\begin{table}[h]
    \centering
    \begin{tabularx}{\textwidth}{@{}lXXX@{}} % {\paperwidth}
        \toprule
        \textbf{Datasets} & \textbf{Train Set} & \textbf{Validation Set} & \textbf{Test Set}\\ 
        \midrule
        FPB & 3100 &  776 & 970\\ 
        \midrule
        FiQA-SA & 750 &  188 & 235\\ 
        \midrule
        Headline-Dir & 6493 & 928  &1856 \\ 
        \midrule
        FinRED & 5655 & 808  & 1616\\ 
        \midrule
        FOMC &1785 & 199  & 496\\ 
        \midrule
        FinArg-AUC-T1 & - & -  & 969\\ 
        \midrule
        M\&A & - &  - & 500\\ 
        \midrule
        FinCausual‘20-T1 & - &  - & 800\\ 
        \bottomrule
    \end{tabularx}
     \caption{Training, validation and test set statistics used for instruction fine-tuning base and instruct models. ‘-’ denotes that the dataset was not used in training.}
    \label{table:4}
\end{table}
%  but we create instructions for its test set. Overall, SOCIALITEINSTRUCTIONS
% contains ~202k data points
% \input{table_models_prompt_templates}

\begin{table}[h]
    \centering
    \label{tab:prompts}
    \begin{tabularx}{\textwidth}{@{}lXX@{}} % {\paperwidth}
        \toprule
        \textbf{Model} & \textbf{Instruct Model Train Prompt Template} & \textbf{Instruct Model Test Prompt Template}\\ 
        \midrule
        Llama3 & \ttfamily <|user|> \{instruction\} \newline sentence: \{input\} <|end|>\newline <|assistant|> \newline label: \{output\} <|end|> & \ttfamily <|user|> \{instruction\} \newline sentence: \{input\} <|end|>\newline <|assistant|> \newline label: \\ 
        \midrule
        Mistral & \ttfamily <s> [INST] <<SYS>> \{instruction\} <</SYS>> \newline Sentence: \{input\} [/INST] \newline {label}: \{output\} </s> & \ttfamily <s> [INST] <<SYS>> \{instruction\} <</SYS>> \newline Sentence: \{input\} [/INST] \newline {label}: \\ 
        \midrule
        Phi-3 & \ttfamily <|begin\_of\_text|><|start\_header\_id|> \newline system<|end\_header\_id|>\{instruction\} <|eot\_id|>\newline <|start\_header\_id|>user<|end\_header\_id|> \newline Sentence:\{input\} <|eot\_id|>  \newline <|start\_header\_id|>assistant<|end\_header\_id|> \newline label: \{output\} <|eot\_id|><|end\_of\_text|> & \ttfamily <|begin\_of\_text|><|start\_header\_id|> \newline system<|end\_header\_id|>\{instruction\} <|eot\_id|>\newline <|start\_header\_id|>user<|end\_header\_id|> \newline Sentence:\{input\} <|eot\_id|>  \newline <|start\_header\_id|>assistant<|end\_header\_id|> \newline label: \\
 
        \bottomrule
    \end{tabularx}
      \caption{Prompt templates for instruct models used in our instruction fine-tuning and test
experiments. Instructions are obtained from Table \ref{table:7}.}
    \label{table:5}
\end{table}

% \begin{table}[ht]
%     \centering
%     \caption{Comparison of Pretrained and Chat Model Prompts}
%     \label{tab:prompts}
%     \begin{tabularx}{\textwidth}{@{}lX@{}}
%         \toprule
%         \textbf{LLM Model} & \textbf{Instruct Model Prompt Template} \\ 
%         \midrule
%         Llama3-8B & \ttfamily <|user|> \{instruction\} \newline sentence: \{input\} <|end|>\newline <|assistant|> label: \{output\} <|end|> \\ 
%         \midrule
%         Mistral-7B & \ttfamily <s> [INST] <<SYS>> \{instruction\} <</SYS>> \newline Sentence: \{input\} [/INST] \newline {label}: \{output\} </s> \\ 
%         \midrule
%         Phi-3-mini & \ttfamily <|begin\_of\_text|><|start\_header\_id|>system<|end\_header\_id|> \newline \{instruction\} <|eot\_id|>\newline <|start\_header\_id|>user<|end\_header\_id|> \newline Sentence:\{input\} <|eot\_id|>  \newline <|start\_header\_id|>assistant<|end\_header\_id|> \newline label: \{output\} <|eot\_id|><|end\_of\_text|>\\
 
%         \bottomrule
%     \end{tabularx}
% \end{table}

\begin{table}[h]
    \centering
    \begin{tabularx}{\textwidth}{@{}lXX@{}} % {\paperwidth}
        \toprule
        \textbf{Base Model Train Prompt Template}\\ 
        \midrule
        \ttfamily Below is an instruction that describes a task, paired with an
        input that provides further context. \\\ttfamily Write a response that
        appropriately completes the request. \\
        \ttfamily\#\#\#Instruction: \\
        \ttfamily\{instruction\} \\
        \ttfamily\#\#\#Input: \\
        \ttfamily\{input\} \\
        \ttfamily\ttfamily\#\#\#Response: \\
        \ttfamily\{output\} \\
        \bottomrule
    \end{tabularx}
     \caption{Prompt templates for base models used in our instruction fine-tuning. Instructions are obtained from Table \ref{table:7}.}
    \label{table:6}
\end{table}

\begin{table}[h]
    \centering
    \label{tab:instruction}
    \begin{tabularx}{\textwidth}{@{}l l X@{}}
        \toprule
        \textbf{Task} & \textbf{Dataset} & \textbf{Instruction} \\ \midrule
        Sentiment analysis & FPB, FiQA-SA & You are a skilled financial analyst specialized in detecting market sentiment from news sources. Your task is to evaluate the sentiment of the following sentence and assign it one of the labels: Positive, Negative, or Neutral. Return only a single word, either Positive or Negative or Neutral. \\ \midrule
        News headline classification & Headline-Dir & You are a skilled financial analyst. Analyze the provided data to identify the trend in price movements of gold. Determine if the price is going up, down, or remaining stable. Return 'up' if the price is increasing, 'down' if it's decreasing, or 'stable' if it's remaining relatively unchanged. Return only a single word, either up or down or stable. \\ \midrule
        Relation extraction & FinRED & You are a skilled financial analyst. Utilize the input text as a context reference, choose the right relationship between 'entity1' and 'entity2' from the options. Return only a single word from the Options. Options: founded by, chief executive officer, employer, product/material produced, industry, owned by, subsidiary, parent organization, manufacturer, brand, owner of, developer, headquarters location, distribution format, original broadcaster, legal form, location of formation, creator, stock exchange, operator, publisher, distributed by, platform, member of, position held, currency, director/manager, chairperson, business division. \\ \midrule
        Hawkish-dovish classification & FOMC & You are an expert financial analyst. Classify the following sentence from FOMC into 'HAWKISH', 'DOVISH', or 'NEUTRAL' class. Label HAWKISH if it is corresponding to tightening of the monetary policy, DOVISH if it is corresponding to easing of the monetary policy, or NEUTRAL if the stance is neutral. \\ \midrule
        Argument unit classification & FinArg-AUC-T1 & You are an expert financial analyst. Analyze sentences from earnings conference calls and identify their argumentative function. Each sentence is either a 'premise', offering evidence or reasoning, or a `claim', asserting a conclusion or viewpoint. Return only a single word, either premise or claim. \\ \midrule
        Deal completeness classification & M\&A & You are an expert financial analyst. In this task, you will be given Mergers and Acquisitions (M\&A) news articles or tweets. Your task is to classify each article or tweet based on whether the mentioned deal was completed or remained a rumour. Your response should be a single word - either `complete' or `rumour' - representing the outcome of the deal mentioned in the provided text. Return only a single word, either complete or rumour. \\ \midrule
        Causal classification & FinCausual'20-T1 & You are an expert financial analyst. In this task, you are provided with sentences extracted from financial news and SEC data. Your goal is to classify each sentence into either `causal' or `noise' based on whether or not it indicates a causal relationship between financial events. Return only a single word, either causal or noise. \\ 
        \bottomrule
    \end{tabularx}
    \caption{Example prompts for each task in our fine-tuning and test experiments}

    \label{table:7}
\end{table}

% \begin{table}[ht]
%     \centering
%     \begin{tabularx}{\textwidth}{X X X c c X}
%         \toprule
%         \textbf{Dataset} & \textbf{Used for} & \textbf{Task} & \textbf{\# Labels} & \textbf{Data size} & \textbf{Example labels} \\ 
%         \midrule
%         FPB & Train, test of ICL, FT & Sentiment analysis & 3 & 4,845 & negative, neutral, positive \\
%         FiQA-SA & Train, test of ICL, FT & Sentiment analysis & 3 & 1,173 & negative, neutral, positive \\
%         Headline-Dir & Train, test of ICL, FT & News headline classification & 3 & 9,277 & up, down, stable \\
%         FinRED & Train, test of ICL, FT & Relation extraction & 29 & 1,070 & employer, industry, owner \\
%         FOMC & Train, test of ICL, FT & Hawkish-dovish classification & 3 & 496 & hawkish, neutral, dovish \\
%         FinArg-AUC-T1 & Evaluation on unseen data & Argument unit classification & 2 & 969 & claim, premise \\
%         M\&A & Evaluation on unseen data & Deal completeness classification & 2 & 500 & complete, rumour\\
%         FinCausual`20-T1 & Evaluation on unseen data & Causal classification & 2 & 8,630 & causal, noise \\ 
%         \bottomrule
%     \end{tabularx}
%     \caption{Datasets}
% \end{table}

% ////////////////////////////////////////Appendix subsection Appendix B ///////////////////////////////////////////////////////////////
\subsection{Appendix B}
\label{subsec:appendix:b}
We present the results of single-task fine-tuning of base and instruct models in Table \ref{table:7} and Figures \ref{fig:7}, \ref{fig:8}, and \ref{fig:9}. The results indicate that single-task fine-tuning performs similarly to multi-task fine-tuning. In some datasets, single-task fine-tuning even relatively outperforms multi-task fine-tuning for base models. For merging the models, we utilized the single-task fine-tuned instruct models, which performed similarly to multi-task fine-tuned models, to alleviate the degradation of zero-shot performance of fine-tuned models on unseen tasks.

%  Our results show that single-task fine-tuning improves the F1 score across all datasets compared to the zero-shot setting and the AdaptLLM-7B baseline model, although it still underperforms the FinMA-7B benchmark on the FPB dataset.

% Notably, we observe substantial improvements, with over a 40\% boost on the FinRed dataset and more than a 20\% boost on the FOMC dataset. This indicates that more challenging tasks, such as predicting monetary policies in the FOMC dataset or identifying relations between entities in the FinRed dataset with its 29-label space, benefit significantly from fine-tuning.
% For simpler tasks such as sentiment analysis and headline classification, the fine-tuned Llama3-8B model slightly outperformed the GPT-4 model. Additionally, the fine-tuned Mistral-7B performed better compared to Llama3-8B and Phi-3 models. This behavior can be attributed to the Instruct fine-tuning process of the Mistral-7B models. This suggests that for single-task fine-tuning, the model size, particularly for models below 10B parameters, is less critical than the quality of the instruction tuning. This is especially relevant since we used instruct models as the starting checkpoints for further fine-tuning.\\

% \input{appendex-result-table}   
% table with acc, f1 and sample data
\begin{table}[ht]
    \centering
    
    \begin{tabular}{@{}ll|ccccccc:ccc@{}}
        \toprule
        \multirow{2}{*}{\textbf{Experiment}} & \multirow{2}{*}{\textbf{Model}} & \multicolumn{2}{c}{\textbf{FPB}} & \multicolumn{2}{c}{\textbf{FiQA-SA}} & \textbf{Headline-Dir} & \textbf{FinRED} & \textbf{FOMC} & \textbf{FinArg} & \textbf{M\&A} & \textbf{SC} \\ 
        \cmidrule(r){3-4} \cmidrule(r){5-6} \cmidrule(r){7-7} \cmidrule(r){8-8} \cmidrule(r){9-9} \cmidrule(r){10-10} \cmidrule(r){11-11} \cmidrule(r){12-12}
        & & \textbf{Acc} & \textbf{F1} & \textbf{Acc} & \textbf{F1} &  \textbf{F1}  & \textbf{F1} & \textbf{F1} & \textbf{F1} &  \textbf{F1} & \textbf{F1} \\ 
        % \midrule
        % \multirow{4}{*}{\textit{Baseline Models}} 
        % & BloombergGPT & - & 51.2 & - & 75.2 & - & - & - & - & - & - \\ 
        % & AdaptLLM-7B & - & 62.5 & - & 72.1 & - & - & - & - & - & - \\ 
        % & FinMA-7B & 86.0 & 86.1 & 78.3 & 79.2 & - & - & 49.0 & 27.5 & 45.3 & - \\ 
        % & GPT-4 & 80.4 & 80.6 & 71.9 & 75.7 & 83.4 & 31.2 & 65.8 & - & - & -  \\ 
        \midrule
        \multirow{3}{*}{\textit{Vanilla Models}} 
        & Llama3-8B & 77.5 & 76.7 & 71.1 & 72.9 & 72.3 & 6.5 & 47.4 & 50.2 & 85.9 & 66.7 \\ 
        & Mistral-7B & 73.4 & 69.8 & 45.9 & 54.8 & 77.6 & 20.5 & 38.6 & 42.2 & 83.6 & 68.8  \\ 
        & Phi-3-mini & 72.8 & 72.9 & 72.7 & 74.8 & 87.1 & 24.8 & 48.5 & 54.1 & 80.1 & 66.5 \\ 
        \midrule
        \multirow{2}{*}{\textit{ST-Base-FT}}
        & Llama3-8B & 84.7 & 84.8 & 88.1 & 88.2 & 95.6 & 74.3 & 67.7 & - & - & - \\ 
        & Mistral-7B & 85.2 & 85.3 & 85.9 & 86.5 & 95.3 & 70.1 & 71.1 & - & - & - \\ 
        % & Phi-3-mini & 83.4 & 82.5 & 76.6 & 79.1 & 95.6 & 69.2 & 66.3 & - & - & -  \\ 
        \midrule
        \multirow{2}{*}{\textit{MT-Base-FT}}
        & Llama3-8B & 79.1 & 79.4 & 87.2 & 85.3 & 95.3 & 69.1 & 68.7 & 23.4 & 72.5 & 52.6 \\ 
        & Mistral-7B & 86.8 & 86.6 & 85.5 & 85.1 & 95.5 & 76.4 & 67.1 & 13.4 & 70.9 & 31.8 \\ 
        % & Phi-3-mini & 83.4 & 82.5 & 76.6 & 79.1 & 95.6 & 69.2 & 66.3 & - & - & -  \\ 
        \midrule
        \multirow{3}{*}{\textit{ST-Instruct-FT}}
        & Llama3-8B & 81.1 & 79.5 & 78.7 & 81.4 & 95.5 & 73.2 & 67.6 & - & - & -   \\ 
        & Mistral-7B & 84.6 & 84.8 & 85.5 & 83.9 & 95.4 & 67.3 & 68.3 & - & - & -  \\ 
        & Phi-3-mini & 83.4 & 82.5 & 76.6 & 79.1 & 95.6 & 69.2 & 66.3 & - & - & -  \\ 
        \midrule
        \multirow{3}{*}{\textit{MT-Instruct-FT}} 
        & Llama3-8B & 86.2 & 86.3 & 86.4 & 86.6 & 95 & 73.4 & 68.4 & 31.1 & 72.9 & 39.2    \\ 
        & Mistral-7B & 86.3 & 86.2 & 85.1 & 83.9 & 95.2 & 69.4 & 70.2 & 35.1 & 76.1 & 57.6   \\ 
        & Phi-3-mini & 84.7 & 84.1 & 79.5 & 81.5 & 95.6 & 67.2 & 66.5 & 53.5 & 77.7 & 64.9 \\
        \midrule
        % \multirow{3}{*}{\textit{Merged Models}} 
        \multirow{2}{*}{\textit{Merged Models}} 
        & Llama3-8B & 80.6 & 80.7 & 75.3 & 78.3 & 93.2 & 44.6 & 60.9 & 54.6 & 74.3 & 65.5  \\ 
        & Mistral-7B & 80.4 & 80.3 & 65.1 & 72.1 & 91.6 & 41.2 & 38.9 & 50.4 & 83.5 & 59.6  \\ 
        % & Phi-3-mini & 40.7 & 87.7 & 24.6 & 84.0 & 16.4 & 81.5 & 14.5 & 70.5 & 18.9 & 30.0  \\
        \bottomrule
    \end{tabular}
    \caption{Main experimental results for four financial classification tasks and three unseen financial classification tasks (FinArg, M\&A, Casual-SC). Vanilla models indicate the zero-shot performance of instruct models. ST-Base-FT refers to single-task fine-tuned base models, and ST-Instruct-FT refers to single-task fine-tuned instruct models. MT-Base-FT refers to multi-task fine-tuned base models, and MT-Instruct-FT refers to multi-task fine-tuned instruct models.}
    \label{table:8}

\end{table}

\begin{figure}[htp]
    \centering
    \includegraphics[width=13cm]{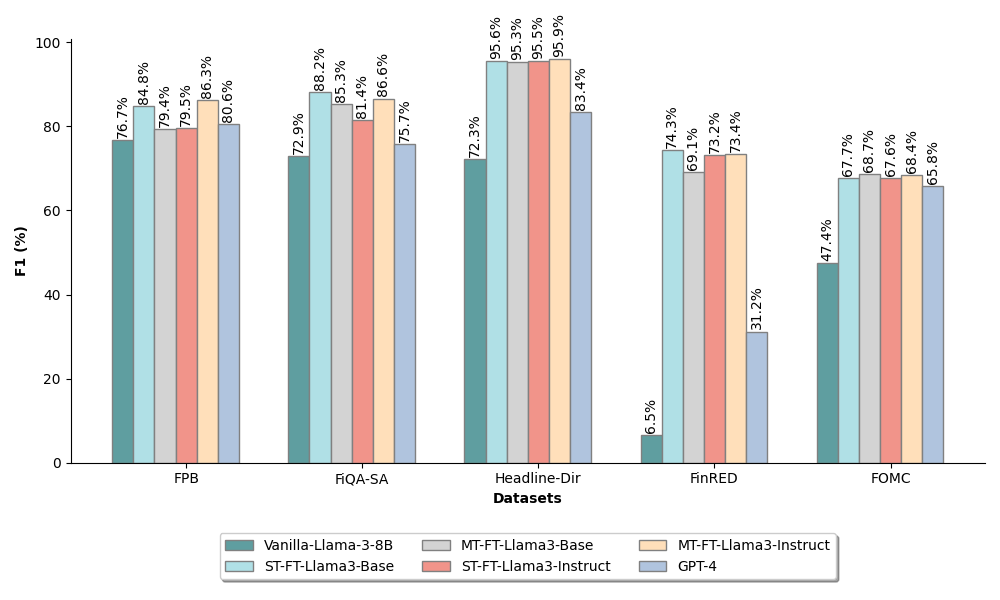}
    \caption{Performance comparison of vanilla models (zero-shot instruct models), single-task fine-tuned base models, multi-task fine-tuned base models, single-task fine-tuned instruct models, and multi-task fine-tuned instruct models on five financial classification datasets for Llama3-8B model. F1 score is reported.}
    \label{fig:7}
\end{figure}

\begin{figure}[htp]
    \centering
    \includegraphics[width=13cm]{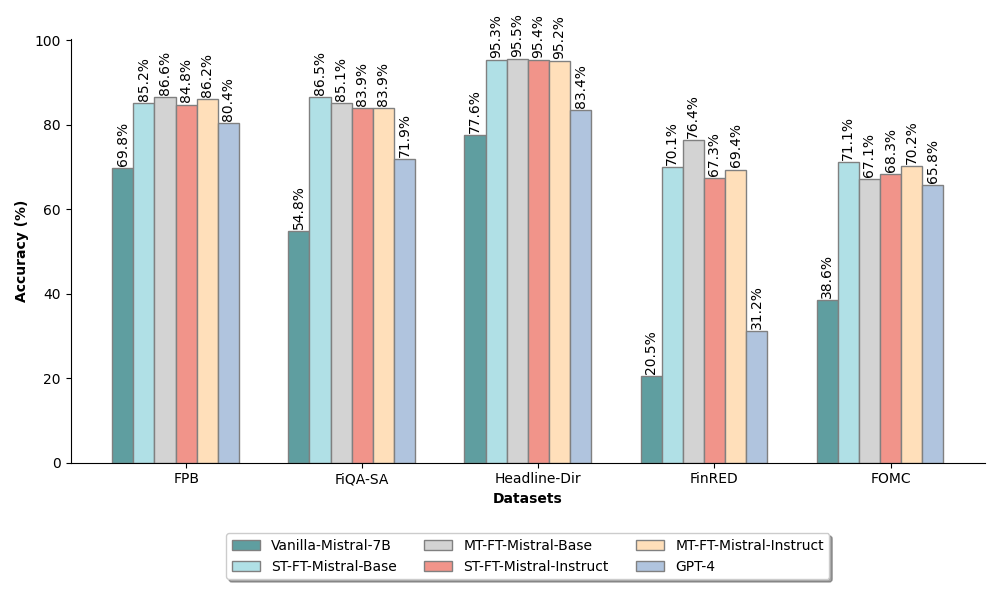}
    \caption{Performance comparison of vanilla models (zero-shot instruct models), single-task fine-tuned base models, multi-task fine-tuned base models, single-task fine-tuned instruct models, and multi-task fine-tuned instruct models on five financial classification datasets for Mistral-7B model. F1 score is reported.}
    \label{fig:8}
\end{figure}
\begin{figure}[htp]
    \centering
    \includegraphics[width=13cm]{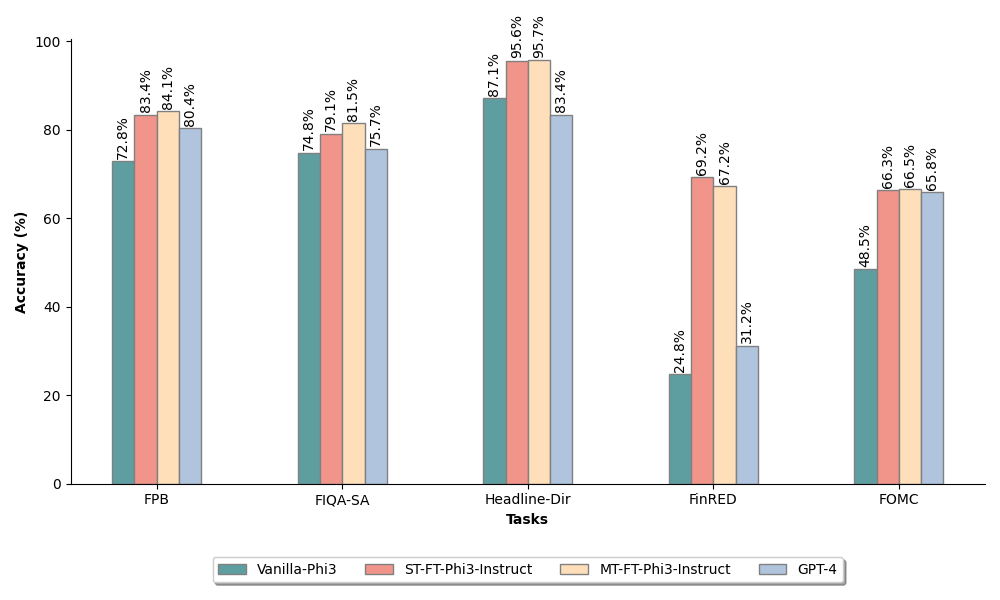}
    \caption{Performance comparison of vanilla models (zero-shot instruct models), single-task fine-tuned instruct models, and multi-task fine-tuned instruct models on five financial classification datasets for Phi-3 model. F1 score is reported.}
    \label{fig:9}
\end{figure}

\end{appendices}
% \begin{figure}[ht] % htbp
%     \centering
%     \input{diagrams/barcharts/fpb_barchart}
%     \caption{Comparison of LLM results for FPB dataset across different experiments.}
%     \label{fig:fpb_barchart_f1scores}
% \end{figure}

% \begin{figure}[h] % htbp
%     \centering
%     \input{diagrams/barcharts/fiqasa_barchart}
%     \caption{Comparison of LLM results for FiQA-SA dataset across different experiments.}
%     \label{fig:fiqasa_barchart_f1scores}
% \end{figure}

% \begin{figure}[h] % htbp
%     \centering
%     \input{diagrams/barcharts/headline_barchart}
%     \caption{Comparison of LLM results for Headline direction dataset across different experiments}
%     \label{fig:headline_barchart_f1scores}
% \end{figure}

% \begin{figure}[h] % htbp
%     \centering
%     \input{diagrams/barcharts/fomc_barchart}
%     \caption{Comparison of LLM results for FOMC dataset across different experiments}
%     \label{fig:fomc_barchart_f1scores}
% \end{figure}

% \begin{figure}[h] % htbp
%     \centering
%     \input{diagrams/barcharts/finred_barchart}
%     \caption{Comparison of LLM results for FinRED dataset across different experiments}
%     \label{fig:finred_barchart_f1scores}
% \end{figure}

\end{document}